\documentclass{article}

\usepackage[main,final]{neurips_2026}

\usepackage[utf8]{inputenc} 
\usepackage[T1]{fontenc}    
\usepackage{hyperref}       
\usepackage{url}            
\usepackage{booktabs}       
\usepackage{amsfonts}       
\usepackage{nicefrac}       
\usepackage{microtype}      
\usepackage{xcolor}         

\usepackage{graphicx}
\usepackage{subcaption}
\usepackage{amsmath}
\usepackage{amssymb}
\usepackage{mathtools}
\usepackage{amsthm}
\usepackage{pifont}
\usepackage{makecell}
\usepackage{multirow}
\usepackage{enumitem}
\usepackage[capitalize,noabbrev]{cleveref}
\usepackage[textsize=tiny]{todonotes}
\usepackage{wrapfig}

\theoremstyle{plain}

\theoremstyle{definition}

\theoremstyle{remark}

\renewcommand{\footnotesize}{\scriptsize}

\title{Fast and Accurate Probing of In-Training LLMs' Downstream Performances}

\author{
    \textbf{Zhichen Liu\textsuperscript{*\,1}},
    \textbf{Tianle Lun\textsuperscript{*\,1}},
    \textbf{Yang Xu\textsuperscript{\ensuremath{\dagger}\,1}}\\
    \textbf{Zhibin Wen\textsuperscript{1}},
    \textbf{Hao An\textsuperscript{1}},
    \textbf{Yulin Ou\textsuperscript{1}},
    \textbf{Jianhui Xu\textsuperscript{1}}\\
    \textbf{Hao Zhang\textsuperscript{2}},
    \textbf{Wenyi Fang\textsuperscript{2}},
    \textbf{Yang Zheng\textsuperscript{2}}
    \\
 \textsuperscript{1} Department of Computer Science and Engineering, \\Southern University of Science and Technology \\
 \textsuperscript{2} Huawei Technologies Co., Ltd. \\
}

\date{}

\begin{document}

\maketitle

\begingroup
\renewcommand{\thefootnote}{\fnsymbol{footnote}}
\footnotetext[1]{Equal contribution.}
\footnotetext[2]{Corresponding author: \texttt{xuyang@sustech.edu.cn}.}
\endgroup




\begin{abstract}
The paradigm of scaling Large Language Models (LLMs) in both parameter size and test time has pushed the boundaries of AI capabilities, but at the cost of making the traditional generative evaluation paradigm prohibitively expensive, therefore making the latency of LLM's in-training downstream performance evaluation unbearable. However, simple metrics like training loss (perplexity) are not always correlated with downstream performance, as sometimes their trends diverge from the actual task outcomes. This dilemma calls for a method that is computationally efficient and sufficiently accurate in measuring model capabilities. 
To address this challenge, we introduce a new in-training evaluation paradigm that uses a lightweight probe for monitoring downstream performance. The probes take the internal representations of LLM checkpoints (during training) as input and directly predict the checkpoint's performance on downstream tasks measured by \emph{success probability} (i.e., pass@1). We design several probe architectures, validating their effectiveness using the OLMo3-7B's, OLMo3-32B's, and Apertus-8B's checkpoints across a diverse set of downstream tasks. The probes can accurately predict a checkpoint's performance (with avg. AUROC$>$0.75), have decent generalizability across checkpoints (earlier predicts later), and reduce the computation latency from $\sim$1 hr (using conventional generative evaluation method) to $\sim$3 min. In sum, this work presents a practical and scalable in-training downstream evaluation paradigm, enabling a more agile, informed, and efficient LLM development process. Our code is available at \url{https://github.com/CLCS-SUSTech/Probe-Prediction}.

\end{abstract}
\section{Introduction}

With the continuous scaling of model parameters \cite{oaiscalinglaw, chinchillascalinglaw}, the engineering behind training Large Language Models (LLMs) has become increasingly complex, necessitating auxiliary mechanisms for in-training supervision. Unlike traditional supervision that prioritizes training stability \cite{palm}, capability-oriented in-training evaluation focuses on verifying whether the capabilities acquired by the model align with expectations for each downstream task \cite{pythia, olmo}. This approach involves tracking evaluation metrics across various downstream tasks rather than relying solely on training perplexity (PPL, i.e., loss). When detecting performance deviations in a specific task, researchers can intervene by adjusting data mixtures to rebalance the model's overall competence \cite{data-mixing-doremi, data-mixing-skillit}. Crucially, perplexity metrics lack this granularity--while perplexity tends to plateau after convergence, performance on downstream tasks often continues to evolve as training progresses \cite{emergent, big-bench}.

As the capability boundaries of LLMs expand, the paradigm of in-training evaluation is shifting from traditional formats (e.g., multiple-choice) toward generative evaluation \cite{helm, openllmleaderboard}. This transition is driven by the predominance of generative demands in modern downstream tasks; generative evaluation not only better reflects LLM's generalizability, but also assesses LLM's ability to handle increasingly complex problems \cite{big-bench, sparkagi}.
However, this expansion aligns with the evolution of scaling laws: while previous scaling focused on parameter count and training data size, recent trends emphasize test-time scaling \cite{test-time-scale}. This shift not only escalates the computational cost of a single forward but also significantly increases the token consumption required for generative evaluation.
The resulting latency poses a severe bottleneck: for instance, if running a full evaluation on a benchmark takes tens of hours, low-latency in-training monitoring becomes computationally prohibitive and practically infeasible.

\begin{figure*}[tbp]
    \centering
    \includegraphics[width=.85\linewidth]{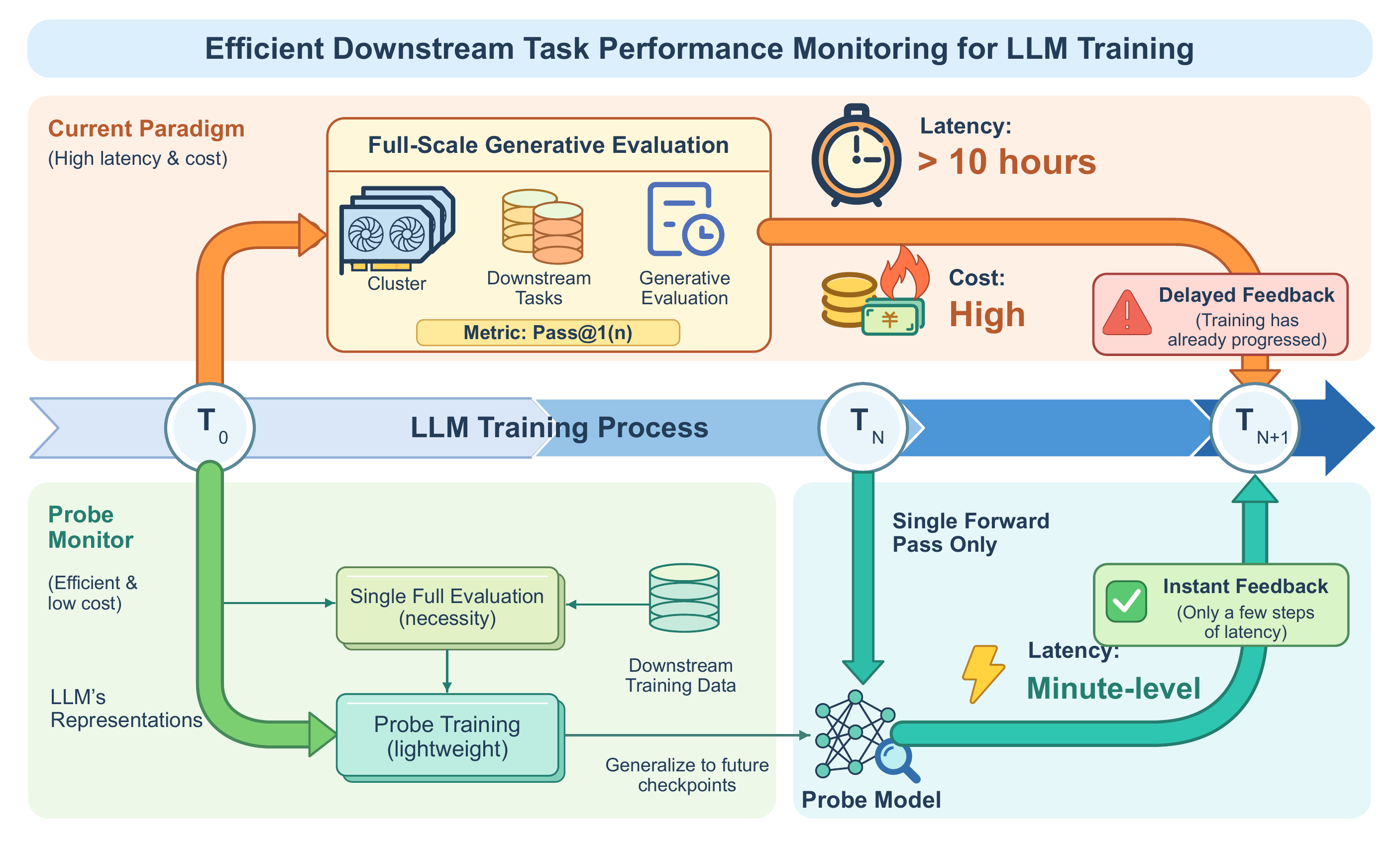}
    \caption{Workflow comparison between the current generative evaluation paradigm and the proposed probing evaluation paradigm for in-training downstream monitoring. Current paradigm suffers from high evaluation cost and excessive latency, resulting in delayed feedback to training progress. In contrast, probe can bypasses the generation process during the evaluation phase, thereby enabling rapid and timely assessment.}
    \label{fig:main}
    \vspace{-1em}
\end{figure*}


For the reasons outlined above, perplexity remains an inadequate surrogate to serve as a reliable proxy for downstream performance. While engineering solutions attempt to reduce evaluation latency by sampling a small-enough subset \cite{tinybench}, this approach inevitably introduces estimation bias \cite{openllmleaderboard}. Consequently, there is an urgent requirement for a new in-training evaluation paradigm that is both accurate and low-latency. To address this challenge, we introduce an efficient probe evaluation framework that utilizes a probe model to monitor downstream performance. The probe model takes the internal states of the model under training and the task input text as inputs, and predicts the corresponding performance metrics (e.g., accuracy and pass@k). Crucially, this allows us to bypass the computationally expensive generation, directly yielding performance predictions.


We investigate several probe architectures, specifically the LoRA-based probe and submodel-based probe, evaluate their performance against some simple baselines. These experiments are conducted under a generative evaluation framework across diverse domains, including general tasks, mathematics, and code generation. The results demonstrate that our method maintains high predictive fidelity (AUROC $>$ 0.75) while drastically reducing evaluation latency, slashing the time required from $\sim$1 hour to $\sim$3 minutes (on OLMo3-7B-base). Consequently, our framework revitalizes the possibility of low-latency, lightweight in-training evaluation once again in the era of modern LLM training.

\begin{figure*}
    \centering
    \includegraphics[width=0.8\linewidth]{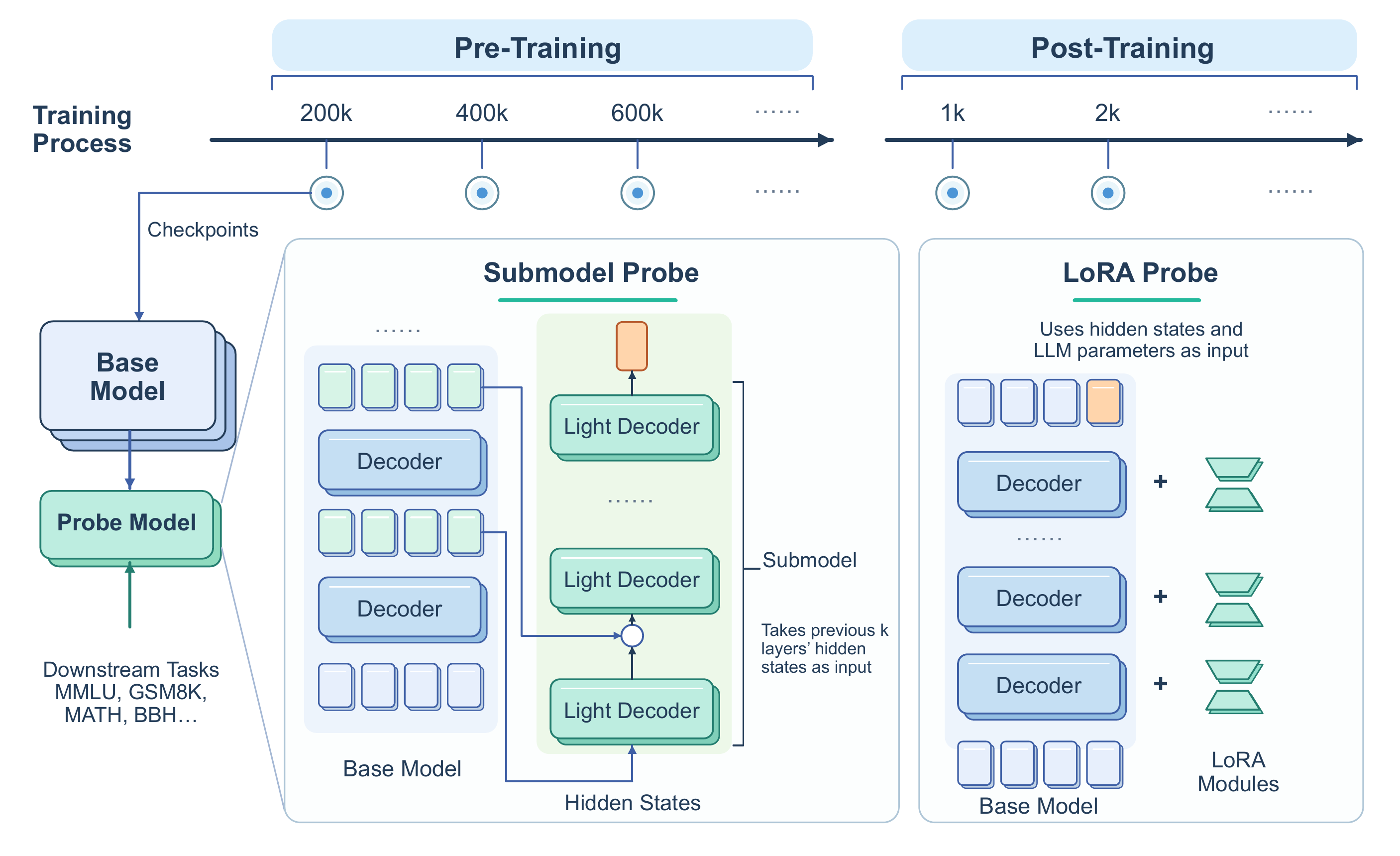}
    \caption{Demonstration of the working process of in-training downstream performance evaluation framework in training stage, and the structure of two lightweight probe models. Training would be applied only in specific checkpoints. Once a probe is trained, it could maintain predictability in the future checkpoints.}
    \label{fig:intro}
    \vspace{-1em}
\end{figure*}

Our main contributions are summarized as follows:
\begin{enumerate}[leftmargin=*, itemsep=0em, topsep=0.em]
    \item[\ding{182}] We propose a novel framework for efficiently in-training downstream evaluation by leveraging lightweight probe models. The proposed method can give feedback accurately with a low latency.
    \item[\ding{183}] We design and evaluate three distinct probe architectures (submodel and LoRA) to predict task-specific success probabilities from the model's internal states.
    \item[\ding{184}] We empirically demonstrate on diverse domains that our trained probes exhibit strong cross-checkpoint generalization, enabling accurate performance prediction for future training steps.
    \item[\ding{185}] Our method provides a practical, low-latency workflow to the current in-training evaluation paradigm, enabling a more agile and informed LLM training process.
\end{enumerate}
\section{Related Works}
\paragraph{Predictability AI} \citet{predictableAI} introduces the fundamental idea of predictable AI that an AI system's key validity indicators can be anticipated (e.g., performance). \citet{analysepredictability} further analyse the predictability of whether the performance of LLM is predictable. Therefore, many works have attempted to better predict the downstream performance of an LLM \cite{downstreamscaling, unveilingdownstreamperformancescaling}. The most popular approach is to train a third-party model for performance prediction \cite{100instanceassessor, proxy, indicators}, while another approach predicts the task-level performance directly rather than individual-level performance \cite{predictingtasklevel}. These works make efforts on LLMs' performance prediction.

\paragraph{LLM In-training Supervision} A vast array of intrinsic metrics inside LLM are proposed to either monitor or stabilize the training process. \citet{attnotallyouneed} explores the rank collapse inside the attention mechanism and uses the relative norm to measure such collapse. \citet{kimik2} discovers the explosion of ``MaxLogit'' and proposes to stabilize training using QK-Clip. Some works find that the massive values inside attention are crucial for contextual understanding \cite{massvalue, attnsink}, while others also find that a stable training process is also related to a stable behavior of attention entropy \cite{attnentropy, attentionentropykeyfactor}. Besides in-training monitoring, \citet{zai-loss} produces a new perspective for in-training evaluation by building the correlation between the loss value and downstream performance.

\paragraph{Probe} 
Recent work suggests there is a ``truthfulness'' direction in latent space~\citep{geometry_of_truth, probing_llm_lying}. Some studies employ linear probes to predict whether a model's answers are correct~\citep{hdndehallu, SEP}. To reduce reliance on the model's generated responses and speed up prediction, certain approaches directly feed the question's hidden state into the probe to assess the model's ability to provide accurate answers~\citep{early_detection, estimateknowledge}. However, correctness and factuality are not entirely linearly encoded~\citep{wehner2025taxonomy}. Therefore, this work proposes the sub-model probe, which captures the dynamic flow of information within the model to predict its performance.
\section{Methods}\label{sec:method}

The primary objective of in-training probe evaluation is to develop a framework for rapidly and efficiently estimating an LLM's performance on downstream tasks without resorting to computationally expensive full-scale evaluations. To achieve this, we first model the generative evaluation process (from input text to generation, then gathering the results to obtain the Pass@1 score) as a regression task, and then introduce a lightweight probe to approximate it.

\subsection{Preliminaries: Performance Estimation as Value Approximation}
The evaluation process can be modeled as value approximation. Let $\mathcal{M}_\theta$ denote a large language model parameterized by weights $\theta$. Given a prompt $x$ from the dataset $\mathcal{D}$, the model generates a response $y$ following the policy $\pi_\theta(y|x)$.
To quantify the performance, we define a ground-truth reward function $R(x, y) \in \{0, 1\}$, which evaluates whether the response $y$ correctly solves the problem in the prompt $x$. Our goal is to estimate the expected performance of $\mathcal{M}_\theta$ on a downstream task without performing a full-scale evaluation.
Adopting a reinforcement learning (RL) perspective, we treat the prompt $x$ as the state $s\in\mathcal{S}$ and the generated response $y$ as the action $a\in\mathcal{A}$. The performance of the model on a specific prompt can be modeled as the state-value function $V^\pi(s)$, which represents the expected reward under the current policy $\pi_\theta$:
\begin{equation}
    V^\pi(s) = \mathbb{E}_{a \sim \pi_\theta(\cdot|s)} [Q^\pi(s, a)] = \sum_{a \in \mathcal{A}} \pi_\theta(a|s) Q^\pi(s, a)
\end{equation}
where $Q^\pi(s, a)$ corresponds to the immediate reward (correctness) of taking action $a$ in state $s$. In our context, since the task is single-turn generation, $Q^\pi(s, a) \equiv R(s, a)$. Thus, predicting the model's accuracy on a prompt is equivalent to approximating the value function $V^\pi(s)$.

\subsection{The Monte Carlo Approximation and Pass@1}
Because the action space $\mathcal{A}$--encompassing all valid token sequences--is theoretically infinite, calculating the exact expectation $V^\pi(s)$ is intractable. We therefore employ a Monte Carlo estimator.
For a given prompt $s$, we sample $n$ independent responses $\{a_1, a_2, \dots, a_n\}$ from the policy $\pi_\theta(\cdot|s)$ to get $\hat{v}(s)$:
\begin{equation}
    \hat{v}(s) = \frac{1}{n} \sum_{i=1}^n R(s, a_i)
\end{equation}
that $\hat{v}(s)$ acts as an approximation of $V^{\pi}(s)$. This $\hat{v}(s)$ serves as the ground truth label for our probe. The objective of our probe $\Phi$ is to minimize the regression loss over the dataset $\mathcal{D}$:
\begin{equation}
    \mathcal{L}(\phi) = \mathbb{E}_{s \sim \mathcal{D}} \left[ || \Phi(s, \theta; \phi) - \hat{v}(s) ||^2 \right]
\end{equation}
where $\phi$ denotes the learnable parameters of the probe.

The widely used $\text{Pass}@1$ metric estimates the probability of correctness when sampling one sample from an $n$-sample set \cite{passk}. From the aspect of $\text{Pass}@1$, $\hat{v}(s)$ is equivalent to $\text{Pass}@1(n)$ when $n$ is held constant. We can analyze this through the definition of $\text{Pass}@1(n)$:
\begin{equation}
\text{Pass}@1(n) = \frac{1}{n} \sum_{i=1}^{n} \mathbb{I}(y_i \text{ is correct})
\end{equation}
which shows the same formulation as $\hat{v}(s)$ ($\mathbb{I}(\cdot)$ is a decision function). Therefore, modeling $\text{Pass}@1(n)$ is equivalent to modeling $V^\pi(s)$. It is worth noting that the commonly used binary metric ``accuracy'' can be viewed as a special case of $\text{Pass}@1$ when $n=1$.

\subsection{Modeling the Probability of Correctness}

\subsubsection{Generative Evaluation}
The ability of an LLM to answer a given question relies on two fundamental sources of information: its internal \textbf{parametric knowledge} ($\theta$), encapsulated within its network weights, and the \textbf{contextual knowledge} ($S$) provided by the input prompt \cite{parametric-knowledge}. The model fuses these two sources to form a high-dimensional latent representation, which we denote as $Rep(S, \theta)\in\mathbb{R}^d$. Subsequently, a generation function, $\psi$, maps this representation to an answer, and a decision function, $\mathbb{I}(\cdot)$, maps the answer to a binary class that represents correct or incorrect. Therefore, the likelihood $p$ of answering the question correctly can be formulated as:
\begin{equation}
    p = \frac{1}{n}\sum^n_{i=1}\mathbb{I}(\psi(Rep(S_i, \theta)) \text{ is correct})
\end{equation}

The entire process is computationally expensive due to the sequential nature of the generation process $\psi$. It is infeasible for frequent evaluation during model training or for large-scale model selection. 

\subsubsection{Probe Evaluation}
To overcome this efficiency bottleneck, we propose replacing the whole process of $\frac{1}{n}\sum^n_{i=1}\mathbb{I}(\psi(\cdot))$ with a lightweight and efficient \textbf{probe model}, denoted as $f(\cdot; \phi)$, where $\phi$ represents the probe's parameters. The probe is trained to directly predict the success probability from the LLM's internal representation:
\begin{equation}
    p \approx \hat{p} = f(Rep(S, \theta); \phi)
\end{equation}
While this approximation may introduce some level of fidelity loss, it provides a sufficiently reliable estimate for monitoring training dynamics and identifying potential issues, at a fraction of the computational cost. The design of the probe hinges on the choice of the representation $Rep(S,\theta)$ which extracts from the base model. We explore two distinct probe architectures based on different hypotheses about where the most salient information resides.

\subsection{Probe Architectures}

In this section, we introduce the architecture of \textbf{Submodel Probe}. The architecture about LoRA Probe is included in Appendix \ref{app:baselines}.

\paragraph{Hypothesis 1} We posit that the model's parametric knowledge is encoded in the progressive transformation of internal states across layers. Therefore, the probe should have access to the hidden states from various depths of the LLM.

This architecture captures parametric knowledge via the base model's hidden states. We postulate that the evolution of hidden states across layers forms an "intrinsic dimension" that correlates with the model's confidence and capability. Let $H^{(l)} \in \mathbb{R}^{T \times d}$ be the hidden states of the base model $\mathcal{M}_\theta$ at layer $l$, where $T$ is the sequence length. We construct a lightweight decoder-only probe model with parameters $\phi$, consisting of $L_{probe}$ layers. The probe's input at layer $k$ integrates its previous state and the projected hidden state from the base model. Formally, let $Z^{(k)} \in \mathbb{R}^{T \times d_{probe}}$ denote the hidden state of the probe at layer $k$. The update rule is:
\begin{equation}
Z^{(k)} = \text{Decoder}^{(k)} \left( Z^{(k-1)} + W_{proj}^{(k)} H^{(k)};\phi^{(k)} \right)
\end{equation}
where $W_{proj}^{(k)} \in \mathbb{R}^{d_{model} \times d_{probe}}$ is a learnable projection matrix, and $k\in K$ matches the probe layer index to the corresponding base model layer index (e.g., $K = N$ for a full-layer mapping, or $K < N$ for an early-layer mapping, $K$ is the number of layers in probe, and $N$ is the number of layers in base model). $H^{(k)} = \text{Decoder}^{(k)}(H^{(k-1)};\theta^{(k)})$ represents the hidden states from base model that contains information from both $S$ and $\theta$.
The final prediction is obtained from the last token representation of the final probe layer $\hat{p}=\sigma(\text{Linear}(Z^{(K)}_{[-1]}))$, where $Z_{[-1]}$ means taking the representation at the last position, and $\sigma$ is the sigmoid function to bound the output within $[0, 1]$. The probe model can be formulated as:
\begin{equation}
\begin{split}
    f_{sub}(Rep(S,\theta);\phi)=&\text{Decoder}^{(k)}\circ \text{Decoder}^{(k-1)} \circ \dots\\ 
    &\circ \text{Decoder}^{(1)}(W_{proj}^{(1)}H^{(1)};\phi)
\end{split}
\end{equation}
where $H^{(1)}=\text{Emb}(X)$ is the output of the embedding layer. Therefore, 
\begin{equation}
    \hat{p}=\sigma(\text{Linear}(f_{sub}(Rep(S,\theta);\phi)_{[-1]}))
\end{equation}

\subsection{In-Training Probe Evaluation Pipeline}
The in-training probe evaluation pipeline requires an initial probe training phase, after which the trained probe can be utilized to evaluate future checkpoints. 
Consider a set of downstream tasks, each containing a training set and a test set (or manually split it if necessary). The pipeline proceeds as follows: 
\begin{enumerate}[leftmargin=*, itemsep=0em, topsep=0.em]
    \item Data Collection: Perform a generative evaluation on the training set to compute the probability of correctness (i.e., Pass@1), which is assigned as the label for each prompt. 
    \item Probe Training: Formulate the problem as a regression task, using the prompt as input and the calculated probability as the training target. 
    \item Probe Evaluation: Feed prompts from the test set into the probe to predict the probability of correctness.
\end{enumerate} 
In terms of computational cost, phases 1 and 2 are the most intensive. However, once a probe is trained, it can predict the Pass@1 value using a single forward pass, significantly accelerating the evaluation process at phase 3.

\section{Experiments}
\label{sec:experiments}

In this section, we empirically validate the effectiveness of the proposed In-Training Probe Evaluation framework. We aim to scrutinize the framework's capability to monitor downstream performance efficiently and accurately throughout the LLM training lifecycle. Specifically, our experiments are designed to address the following research questions:
\begin{itemize}[leftmargin=*, itemsep=0pt, topsep=0pt]
    \item \textbf{RQ1 (Fidelity):} Can lightweight probes accurately approximate the downstream performance of LLMs solely based on internal representations?
    \item \textbf{RQ2 (Generalizability):} Do probes trained on early checkpoints exhibit transferability to future checkpoints, despite the significant distribution shifts incurred during training?
    \item \textbf{RQ3 (Efficiency):} What is the magnitude of computational acceleration achieved by our framework compared to traditional generative evaluation paradigms?
    \item \textbf{RQ4 (Robustness under Controlled Degradation)}: Do probes still maintain effectiveness on unstable checkpoints?
\end{itemize}

\subsection{Experimental Setup}

\paragraph{Datasets Preparation.}
To ensure a comprehensive assessment of model capabilities, we curate a diverse benchmark suite spanning three primary domains:
(1) \textbf{Reasoning}: We employ GSM8K \cite{gsm8k}, MATH \cite{math}, AIME \cite{aime1983, aime25} to evaluate mathematical problem-solving, GPQA \cite{gpqa} for graduate-level scientific reasoning, and BBH \cite{bbh} for symbolic reasoning tasks.
(2) \textbf{Knowledge}: The MMLU \cite{mmlu} benchmark is utilized to assess massive multitask language understanding.
(3) \textbf{Coding}: We adopt HumanEval~\cite{humaneval} and MBPP~\cite{mbpp} to evaluate code generation proficiency.

For each dataset, we construct the training targets by performing a standard generative evaluation. Specifically, we define the ground truth label as the empirical success probability, estimated by sampling $n$ responses per prompt (denoted as Pass@1($n$)). Detailed statistics and configuration of the datasets are summarized in \Cref{tab:datainfo}. Further implementation details are provided in Appendix~\ref{app:data_preparation}.

\begin{table}[tbp]
    \centering
    \caption{Summary of datasets used for probe training and evaluation. \textbf{N-shot} denotes the number of in-context exemplars. \textbf{Stage} indicates the applicable training phase: \textit{Pre} (Pre-training) and \textit{Post} (Post-training/Instruction Tuning). \textbf{Sampling Budget ($n$)} refers to the number of samples generated to estimate the ground-truth Pass@1 score in pre/post stage.}
    \label{tab:datainfo}
    \resizebox{0.8\linewidth}{!}{
    \begin{tabular}{l|cccccc}
    \toprule
    \textbf{Dataset} & \textbf{Domain} & \textbf{N-shot} & \textbf{Stage} & \textbf{\makecell{Sampling\\Budget}} & \textbf{Train Size} & \textbf{Test Size} \\
    \midrule
    MMLU & Knowledge & 5 & Pre/Post & 8/4 & 10k & 14k \\
    GSM8K & Math & 8 & Pre/Post & 8/4 & 7.5k & 1.3k \\
    MATH & Math & 4 & Pre/Post & 8/4 & 6k & 500 \\
    BBH & Reasoning & 3 & Pre/Post & 8/4 & 5.2k & 1.4k \\
    HumanEval & Code & 0 & Pre/Post & 8/4 & 100 & 64 \\
    MBPP & Code & 3 & Pre/Post & 8/4 & 710 & 250 \\
    AIME & Math & 0 & Post & -/16 & 933 & 60 \\
    GPQA & Science & 0 & Post & -/8 & 350 & 98 \\
    \bottomrule
    \end{tabular}
    }
\end{table}


\begin{table*}[tbp]
    \centering
    \caption{\textbf{Estimation fidelity on Pre-training Checkpoints.} The models are trained and evaluated on the OLMo-3-Base-800k checkpoint. ($\uparrow$: higher is better; $\downarrow$: lower is better).}
    \resizebox{0.8\linewidth}{!}{
    \begin{tabular}{lc|cccccc|c}
    \toprule
        & Metric & MMLU & GSM8k & MATH & BBH & HumanEval & MBPP & Avg.\\
    \midrule
    \multirow{2}{*}{Loss fit} & AUROC ($\uparrow$) &0.4588 & 0.6189 & 0.6177 & 0.5219 & 0.7048 & 0.5549 & 0.5795\\
    & MSE ($\downarrow$) & 0.2322 & 0.1239 & 0.0399 & 0.1880 & 0.1407 & 0.1755 & 0.1500\\
    \midrule
    \multirow{2}{*}{Linear} & AUROC ($\uparrow$) & 0.6019 & 0.5036 & 0.4714 & 0.7309 & 0.5966 & 0.5818 & 0.5810\\
    & MSE ($\downarrow$) & 0.1780 & 0.1310 & 0.0406 & 0.1534 & 0.1773 & 0.1686 & 0.1415\\
    \midrule
    \multirow{2}{*}{Submodel} & AUROC ($\uparrow$) & \textbf{0.6737} & \textbf{0.7827} & \textbf{0.7776} & \textbf{0.8014} & \textbf{0.9163} & \textbf{0.7822} & \textbf{0.7890}\\
    & MSE ($\downarrow$) & \textbf{0.1492} & \textbf{0.0962} & \textbf{0.0333} & \textbf{0.1302} & \textbf{0.0810} & \textbf{0.1395} & \textbf{0.1049} \\
    \midrule
    \multirow{2}{*}{LoRA} & AUROC ($\uparrow$) & 0.5519 & 0.6737 & 0.6983 & 0.7731 & 0.6565 & 0.7031 & 0.6761\\
    & MSE ($\downarrow$) & 0.2049 & 0.1201 & 0.0436 & 0.1399 & 0.1463 & 0.1501 & 0.1342 \\
    \bottomrule
    \end{tabular}
    }
    \label{tab:avg_auroc}
\end{table*}

\paragraph{Base Models and Checkpoints.}
We conduct our experiments mainly using the OLMo3-7B's and OLMo3-32B's checkpoints \cite{olmo3}, because of its open accessibility to granular checkpoints across the training trajectory (the results of 32B model are demonstrated in Appendix \ref{app:result-olmo3-32b}). In addition, we also conduct experiments on Apertus-8B\cite{swissai2025apertus} to test the transferability of our method. To rigorously evaluate the robustness of our probes against distribution shifts, we utilize a sequence of checkpoints spanning two distinct training phases: 
\begin{enumerate}[leftmargin=*, itemsep=0em, topsep=0em] 
\item \textbf{Pre-training Phase:} We sample intermediate checkpoints at varying step intervals during the fundamental language modeling stage (i.e., OLMo-3-1025-7B\footnote{\url{https://huggingface.co/allenai/Olmo-3-1025-7B}} and Apertus-8B-2509\footnote{https://huggingface.co/swiss-ai/Apertus-8B-2509}). 
\item \textbf{Post-training Phase:} We include checkpoints from instruction tuning and alignment stages (i.e., OLMo-3-7B-Instruct\footnote{\url{https://huggingface.co/allenai/Olmo-3-7B-Instruct}} and OLMo-3-7B-Think\footnote{\url{https://huggingface.co/allenai/Olmo-3-7B-Think}}). 
\end{enumerate}

\paragraph{Baselines}
To benchmark the effectiveness of our proposed architectures, we compare them against 
two established baselines: 1. \textbf{Loss fit}, a conventional metric that uses the negative log-likelihood to predict the value. This baseline assumes that lower perplexity correlates with higher downstream accuracy. 2. \textbf{Linear probe}, a simplified version of our framework, where each layer trained a linear regressor to predict the final value. This serves to isolate the gains attributed to our proposed non-linear probe architectures (Submodel and LoRA). See Appendix \ref{app:baselines} for more implementation details. We also provide a comparison with subset sampling in Appendix \ref{app:subset_sampling}.

\paragraph{Evaluation Metrics.}
The probes are trained to regress the success probability $\hat{p}$. To quantify the estimation quality, we applied a binary classification process to the Pass@1 value in metrics computation: We binarized the ground truth Pass@1 values using a threshold of 0.5, i.e., $y = \mathbb{I}(\text{Pass@1} \geq 0.5)$ and report the \textbf{Area Under the Receiver Operating Characteristic Curve (AUROC)}. AUROC is threshold-independent and robust to class imbalance, making it an ideal metric to measure how well the probe's predicted scores rank successful instances higher than failed ones. A larger AUROC value means better probe performance:  $\text{AUROC}=1.0$ indicates perfect prediction; $\text{AUROC}=0.5$ indicates random guessing. Additionally, we also reported Mean Squared Error (MSE) $\lVert \hat{p}-p \rVert_2$ as an evidence for goodness of fit.

\begin{table*}[tbp]
  \centering
  \vspace{-1em}
  \caption{\textbf{Forward Transferability Analysis.} We report AUROC scores where probes are trained on earlier checkpoints (rows) and tested on future checkpoints (columns).}
\resizebox{0.8\linewidth}{!}{
  \begin{tabular}{c|c|c|c|c|c|c|c|c|c}
    \toprule
    & \multirow{2}{*}{\makecell{Train\\Ckpt (k)}} & \multicolumn{8}{c}{\textbf{Test Ckpt (k)}}\\
    \cline{3-10}
    & & \textbf{200} & \textbf{400} & \textbf{600} & \textbf{800} & \textbf{1000} & \textbf{1200} & \textbf{1400} & \textbf{last} \\
    \midrule
    \multirow{7}{*}{Submodel}& 200  & \textbf{0.7639} & 0.7346 & 0.7523 & 0.7633 & 0.7504 & 0.7379 & 0.7414 & 0.7513 \\
    & 400  & {--}   & 0.7491 & 0.7654 & \textbf{0.7764} & 0.7647 & 0.7561 & 0.7554 & 0.7657 \\
    & 600  & {--}   & {--}   & 0.7773 & \textbf{0.7850} & 0.7697 & 0.7617 & 0.7628 & 0.7685\\
    & 800  & {--}   & {--}   & {--}   & \textbf{0.7890} & 0.7719 & 0.7695 & 0.7680 & 0.7731\\
    & 1000 & {--}   & {--}   & {--}   & {--}   & \textbf{0.7728} & 0.7608 & 0.7630 & 0.7687\\
    & 1200 & {--}   & {--}   & {--}   & {--}   & {--}   & 0.7667 & 0.7655 & \textbf{0.7741}\\
    & 1400 & {--}   & {--}   & {--}   & {--}   & {--}   & {--}   & 0.7709 & \textbf{0.7780}\\
    \midrule
    \multirow{7}{*}{LoRA}& 200  & \textbf{0.6741} & 0.6318 & 0.6457 & 0.6132 & 0.5958 & 0.5728 & 0.5831 & 0.5683 \\
    & 400  & {--}   & \textbf{0.6729} & 0.6733 & 0.6477 & 0.5949 & 0.5741 & 0.6028 & 0.5981\\
    & 600  & {--}   & {--}   & \textbf{0.7097} & 0.6574 & 0.6179 & 0.5964 & 0.6284 & 0.6115\\
    & 800  & {--}   & {--}   & {--}   & \textbf{0.6761} & 0.6245 & 0.5891 & 0.6434 & 0.6269\\
    & 1000 & {--}   & {--}   & {--}   & {--}   & \textbf{0.6432} & 0.5993 & 0.6313 & 0.6126\\
    & 1200 & {--}   & {--}   & {--}   & {--}   & {--}   & \textbf{0.6744} & 0.6607 & 0.6433\\
    & 1400 & {--}   & {--}   & {--}   & {--}   & {--}   & {--}   & \textbf{0.6744} & 0.6318\\
    \bottomrule
  \end{tabular}
  }
  \vspace{-0.5em}
  \label{tab:olmo3-base-auroc}
\end{table*}

\subsection{RQ1: High-Fidelity Performance Estimation}

We first assess the fundamental capability of the proposed probes to approximate ground-truth performance metrics (Pass@1) using only internal representations. \Cref{tab:avg_auroc} presents the comparative results on the pre-training checkpoints. 
The Loss Fit baseline yields an average AUROC of only 0.5795, barely outperforming random guessing (0.5). This corroborates the ``emergence'' phenomenon in LLMs, where training loss cannot correctly correlate with the ability to solve complex reasoning tasks. Similarly, the Linear probe fails to extract sufficient signals (Avg. AUROC of 0.5810), suggesting that the mapping from static representations to task success is highly non-linear. 
The proposed Submodel architecture significantly outperforms all baselines, achieving an average AUROC of 0.7890 and reducing MSE to 0.1049. From these results, we can conclude that Submodel is sufficient to approximate the correctness with high relevance and low error. 

Notably, probes on tasks like HumanEval and BBH can achieve higher AUROCs compared to other tasks, while MMLU is less predictable. The  discussion and more related information are in Appendix \ref{app:results}, where we demonstrate all results in the pre-training and post-training stages.

\subsection{RQ2: Cross-Checkpoint Generalizability}
A critical requirement for in-training monitoring is \textbf{Forward Transferability}: can a probe trained on an earlier checkpoint ($t$) accurately forecast the performance of a future checkpoint ($t + \Delta t$)? We investigate this by training probes on specific checkpoints (rows in \Cref{tab:olmo3-base-auroc}) and testing them on subsequent checkpoints (columns). 

\begin{wrapfigure}{R}{0.45\textwidth}
    \centering
    \includegraphics[width=.9\linewidth]{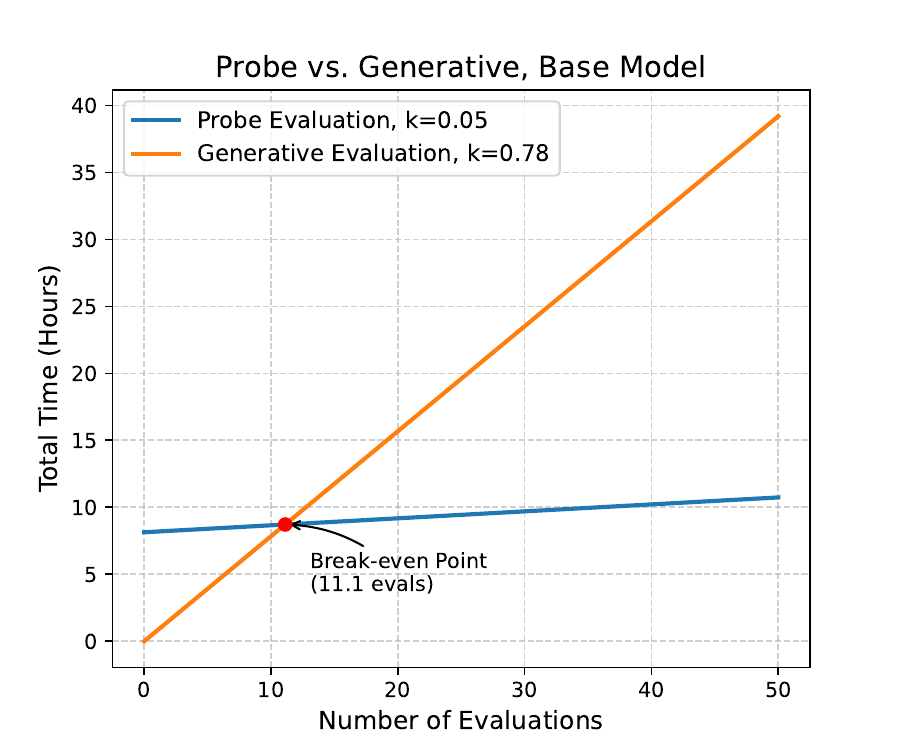}
    \caption{Cumulative time consumption for Probe Evaluation vs. Generative Evaluation on pre-training checkpoints (OLMo-3-Base).}
    \label{fig:base_time}
\end{wrapfigure}

As shown in \Cref{tab:olmo3-base-auroc}, the Submodel probe demonstrates exceptional temporal robustness. For instance, a probe trained at step 200k maintains an AUROC of $\sim0.75$ even when evaluated on the final checkpoint (Last), despite massive updates to the base model's weights. This implies that the mapping between hidden state dynamics and task correctness is an \emph{invariant property} of the model architecture, independent of specific weight values. 

Conversely, the LoRA probe suffers from significant degradation when transferred to future checkpoints (e.g., dropping from 0.67 to 0.56). This confirms our hypothesis: LoRA learns to modulate specific weight matrices ($W_\theta$), and as $W_\theta$ drifts during training, the learned modulation $\Delta W_\phi$ becomes obsolete. In contrast, the Submodel operates on the \emph{semantic representation space}, which evolves more smoothly. 

\begin{table*}[tbp]
    \centering
    \caption{\textbf{Forward Transferability on Apertus-8B-2509.} AUROC scores for Submodel probes trained on earlier checkpoints (rows) and tested on later checkpoints (columns).}
    \resizebox{0.8\linewidth}{!}{
    \begin{tabular}{c|c|c|c|c|c|c|c|c}
    \toprule
    & \multirow{2}{*}{\makecell{Train\\Ckpt (k)}} & \multicolumn{7}{c}{\textbf{Test Ckpt (k)}} \\
    \cline{3-9}
    & & \textbf{400} & \textbf{800} & \textbf{1194} & \textbf{1670} & \textbf{2000} & \textbf{2400} & \textbf{2627} \\
    \midrule
    \multirow{7}{*}{Submodel}
    & 400  & \textbf{0.7812} & 0.7163 & 0.7384 & 0.7381 & 0.7482 & 0.7471 & 0.7425 \\
    & 800  & -- & \textbf{0.7265} & 0.7413 & 0.7331 & 0.7549 & 0.7417 & 0.7602 \\
    & 1194 & -- & -- & \textbf{0.7442} & 0.7410 & 0.7541 & 0.7464 & 0.7500 \\
    & 1670 & -- & -- & -- & \textbf{0.7511} & 0.7613 & 0.7542 & 0.7614 \\
    & 2000 & -- & -- & -- & -- & \textbf{0.7707} & 0.7557 & 0.7656 \\
    & 2400 & -- & -- & -- & -- & -- & \textbf{0.7706} & 0.7677 \\
    & 2627 & -- & -- & -- & -- & -- & -- & \textbf{0.7859} \\
    \bottomrule
    \end{tabular}
    }
    \label{tab:apertus-auroc}
\end{table*}

\paragraph{Transferability on Apertus-8B.}
To test transferability beyond the OLMo3 model family, we perform the same experiment on Apertus-8B-2509 checkpoints. We train Submodel probes on seven pre-training checkpoints from 400k to 2627k steps and test on the same six benchmarks as Olmo3-7B. As shown in \Cref{tab:olmo3-base-auroc}, probes trained on early checkpoints retain AUROCs of 0.716--0.768 on later checkpoints. In particular, the 400k probe obtains 0.7425 AUROC on the final 2627k checkpoint, and the mean over all results is 0.7485. These results extend the evidence beyond the OLMo3 trajectory. Complete MSE results are provided in Appendix~\ref{app:apertus-mse}.


\subsection{RQ3: Computational Efficiency and Amortized Cost}

While the accuracy and generalizability have been demonstrated, another core focus of the proposed framework is efficiency, that is, whether it can replace traditional generative evaluation to achieve low-latency assessment.
We analyze the computational cost of the probe evaluation pipeline compared to standard generative evaluation. The total time cost $T$ can be modeled as $T = T_{init} + N \times t_{eval}$, where $N$ is the number of evaluations. Generative evaluation does not contain a $T_{init}$, however, it usually contains a high $t_{eval}$. While probe evaluation requires a preparation phase, its $t_{eval}$ is significantly smaller than generative evaluation. 
As illustrated in \Cref{fig:base_time}, although the probe method requires an initialization phase (data collection + training), it achieves a \textbf{15.6$\times$ speedup} in inference per checkpoint on the base model--while generative evaluation requires 0.78h to finish an evaluation, probe evaluation only takes 0.05h. The advantage becomes even more pronounced in post-training stages where generation length increases. As detailed in Appendix~\ref{app:time}, the speedup factors reach \textbf{61.4$\times$} for instruct models and \textbf{231.8$\times$} for think models. This makes our framework particularly suitable for high-frequency monitoring of modern reasoning models.

\subsection{RQ4: Robustness under Controlled Degradation}

The released checkpoints used above are mainly on stable training trajectories. To test the robustness of our probes on unstable checkpoints, we construct two degraded variants of the OLMo3-7B 800k checkpoint by adding zero-mean Gaussian noise to each parameter with relative scales of 0.1 and 0.2, where the standard deviation of the noise for each parameter is proportional to its RMS.

\begin{table}[tbp]
    \centering
    \caption{\textbf{Controlled degradation test.} Scores are pooled averages over all evaluation samples from the six benchmarks.}
    \label{tab:olmo3-stress}
    \resizebox{0.6\linewidth}{!}{
    \begin{tabular}{l|cc}
    \toprule
    Checkpoint & Actual average & Predicted average \\
    \midrule
    OLMo3-800k (original) & 0.4977 & 0.5159 \\
    OLMo3-800k + noise 0.1 & 0.4585 & 0.5036 \\
    OLMo3-800k + noise 0.2 & 0.2923 & 0.4556 \\
    \bottomrule
    \end{tabular}
    }
\end{table}

\begin{table}[tbp]
    \centering
    \caption{\textbf{AUROC under controlled degradation.} AUROC is computed over all six benchmarks.}
    \label{tab:olmo3-stress-auroc}
    \resizebox{0.8\linewidth}{!}{
    \begin{tabular}{l|cccccc}
    \toprule
    Probe & 200k & 400k & 600k & 800k & 800k + noise 0.1 & 800k + noise 0.2 \\
    \midrule
    200k & 0.7639 & 0.7346 & 0.7523 & 0.7633 & 0.7566 & 0.7511 \\
    400k & -- & 0.7491 & 0.7654 & 0.7764 & 0.7660 & 0.7625 \\
    600k & -- & -- & 0.7773 & 0.7850 & 0.7754 & 0.7526 \\
    800k & -- & -- & -- & 0.7890 & 0.7758 & 0.7603 \\
    \bottomrule
    \end{tabular}
    }
\end{table}

As shown in \Cref{tab:olmo3-stress}, the actual pooled score decreases from 0.4977 to 0.4585 and 0.2923 as the noise level increases. The probe also predicts a decrease, from 0.5159 to 0.5036 and 0.4556, and correctly identifies the stronger degradation under noise level 0.2, although it underestimates the magnitude of drop. 
As shown in \Cref{tab:olmo3-stress-auroc}, across probes trained on 200k, 400k, 600k, and 800k, AUROC remains in the ranges 0.757--0.776 and 0.751--0.762 under noise levels 0.1 and 0.2, respectively. We view these results as evidence that the probe can serve as a low-cost, high-frequency warning signal for substantial model degradation.

\subsection{Ablation: Layer-wise Information Sufficiency}
\begin{table*}[tbp]
    \centering
    \caption{\textbf{Impact of Input Layer Depth.} Comparison of Submodel probes trained on the first $k$ layers versus the full model (32 layers).}
    \resizebox{0.8\linewidth}{!}{
    \begin{tabular}{c|cccccc|c}
    \toprule
    Num of Layers & MMLU & GSM8k  & MATH & BBH & HumanEval & MBPP & Avg. \\
    \midrule
    8-layer       & 0.6079 & 0.6663 & 0.6921 & 0.7894  & 0.8612  &  0.5911 & 0.7013\\
    16-layer & 0.6937 & 0.7463 & 0.7636 & 0.7963  &  0.9184 & 0.7166 & 0.7725 \\
    \midrule
    Full (32-layer)      & 0.6737 & 0.7827 & 0.7776 & 0.8014 & 0.9136 & 0.7822 & 0.7890\\
    \bottomrule
    \end{tabular}
    }
    \label{tab:submodel-layer}
\end{table*}

Finally, we investigate the minimal information required for accurate prediction. The Submodel probe normally aggregates hidden states from all layers. Here, we analyze the performance impact of restricting the input to the first $k$ layers (``Early Exit'' strategy). 
\Cref{tab:submodel-layer} reveals that the first 16 layers contain the majority of the signal required for performance estimation (Avg. AUROC 0.7725 vs. 0.7890). However, reducing the scope to only the first 8 layers results in a notable performance drop, indicating that critical reasoning cues are distributed across the mid-to-upper layers of the network. This finding supports the feasibility of further accelerating the pipeline by caching only the upper-layer representations.
\section{Conclusion}

In this work, we address the critical bottleneck of efficiently monitoring the downstream performance of LLMs during training, where the imperative for frequent monitoring conflicts with the computationally prohibitive cost of standard generative benchmarks. We introduce the In-Training \textbf{Probe Evaluation} framework, a novel paradigm that substitutes the high latency generative evaluation paradigm with agile, efficient diagnostic probes. By formulating performance estimation as a value approximation problem, we demonstrate that lightweight probes can accurately regress the probabilities of correctness on downstream tasks from the model's internal representations, bypassing the need for full-scale generation.

Our extensive experiments yield two pivotal insights. First, we establish that the correlation between internal hidden states and task correctness is remarkably invariant to the distribution shifts inherent in model training. Specifically, our Submodel probe architecture exhibits strong \textit{forward transferability}: a probe trained on previous checkpoints can reliably forecast the performance of future checkpoints. Second, we quantify the efficiency gains, demonstrating that our framework reduces evaluation latency by orders of magnitude ($15.6\times$ speed up compared with generative evaluation), rendering high-frequency, granular monitoring feasible once again.

Looking ahead, this work lays the foundation for a more controllable and scientifically grounded LLM development process. The implications extend beyond passive monitoring; our framework offers a viable path toward active training intervention. The high-fidelity signals provided by these probes could serve as real-time feedback mechanisms for dynamic data curriculum learning—allowing developers to adjust data mixtures or hyperparameters on-the-fly to optimize specific downstream capabilities. By decoupling evaluation costs from model scale, we provide a scalable methodology to accelerate the iteration and understanding of next-generation AI systems.

\newpage

\section*{Acknowledge}
We sincerely thank all reviewers for giving constructive feedbacks that make this paper much improved. 
This study is funded by Shenzhen Science and Technology Program (No. JCYJ20240813094612017), Guangdong Province ZJRC Program (No. 2024QN11X145), and Huawei Technologies Co., Ltd. (No. TC20211206598-2024-06). 

\bibliographystyle{icml2026}

\bibliography{main}

\newpage

\appendix
\onecolumn

\section{Data Preparation}
\label{app:data_preparation}

\subsection{Preparation Process}
Data preparation requires splitting datasets into training and test sets. For MMLU and GSM8k, which already contain standard splits, we use them directly. However, other datasets do not provide a training split; therefore, we manually perform a train-test split for BBH, HumanEval, MBPP, and GPQA. Specifically, for MATH, we use MATH-500 \cite{math-500} as the test set and the original MATH dataset (excluding MATH-500) as the training set. For BBH, which consists of multiple subsets, we select 50 samples from each subset for the test set and use the remaining data for training. Finally, we randomly downsample the training sets of MMLU and MATH, as their large size would make the evaluation time cost prohibitive.

\subsection{MMLU Prompt Format}

\texttt{Think step by step and answer the following questions. \\ \\
Question:~\{question\} \\
(A)~\{option 1\} (B)~\{option 2\} (C)~\{option 3\} (D)~\{option 4\} \\ \\
Answer:~\{answer\} \\ \\
\{the rest 4 shots\} \\ \\
Question:~\{question\} \\ \\
Answer: Let’s think step by step.
}

\subsection{GSM8K Prompt Format}

\texttt{Question:~\{question\} \\
Answer:~\{chain-of-thought\} \\
\#\#\#~\{answer\} \\ \\
\{the rest 7 shots\} \\ \\
Question:~\{question\} \\
Answer:
}

\subsection{MATH Prompt Format}

\texttt{Please reason step by step, and put your final answer within \textbackslash boxed\{\}. \\ \\
Question:~\{question\} \\ \\
Answer:~\{answer\} \\ \\
\{the rest 3 shots\} \\ \\
Question:~\{question\} \\ \\
Answer: 
}

\subsection{BBH Prompt Format}

\texttt{\{task description\} \\ \\
QUESTION:~\{question\} \\
OPTIONS: \\~\{options\} \\
Let's think step by step. \\
\{chain-of-thought\} \\
ANSWER:~\{answer\} \\ \\
\{the rest 2 shots\} \\ \\
QUESTION:~\{question\} \\
OPTIONS:~\{options\} \\
Let's think step by step.
}

\subsection{MBPP Prompt Format}

\texttt{You are an expert Python programmer, and here is your task:~\{task\}.Your code should pass these tests:\\ \\
\{tests\} \\ 
{}[BEGIN] \\
\{code\} \\
{}[DONE] \\
\{the rest 2 shots\} \\
You are an expert Python programmer, and here is your task:~\{task\}.Your code should pass these tests:\\ \\
\{tests\} \\ 
{}[BEGIN] \\
}

\subsection{AIME Prompt Format}

\texttt{Please reason step by step, and put your final answer within \textbackslash boxed\{\}.\\ \\
\{question\}
}

\subsection{GPQA Prompt Format}

\texttt{Please reason step by step, and put your final answer within \textbackslash boxed\{\}.\\ \\
Question:\\\{question\}\\ \\
Answer:
}

\section{Baselines}
\label{app:baselines}

\subsection{LoRA Probe}

\paragraph{Hypothesis 2} We hypothesize that the most direct access to task-specific parametric knowledge is achieved through lightweight perturbations of the frozen model weights. 

This architecture captures parametric knowledge by directly interacting with the weights via Low-Rank Adaptation (LoRA) \cite{lora}. We interpret the LoRA mechanism not just as fine-tuning, but as a probe that modulates the frozen parametric knowledge $\theta$ based on contextual knowledge $S$. 
If we treat the LoRA module $AB$ as the probe model's weight $W_{\phi}$, the standard LoRA formulation can be rewritten as:
\begin{equation}
    Y = (W_\theta + AB)X = W_{\phi}X + W_\theta X
\end{equation}
therefore the probe model takes both $X \in S$ and $W_\theta \in \theta$ as inputs through the following formulation:
\begin{equation}
\begin{split}
    f_{lora}(Rep(S,\theta); \phi) = &\tilde{D}^{(n)} \circ \tilde{D}^{(n-1)} \circ \dots \\&\circ \tilde{D}^{(1)}(H^{(1)};\theta,\phi)
\end{split}
\end{equation}
where $\tilde{D}(H;\theta,\phi)$ is a standard decoder layer operated by LoRA through $\tilde{W}=W_\theta+AB$. We attach LoRA adapters $\phi = \{A_l, B_l\}_{l=1}^L$ to the base model. Unlike standard LoRA for text generation, we optimize these adapters solely for the regression task:
\begin{equation}
    \hat{p}=\sigma(\text{Linear}(f_{lora}(Rep(S,\theta);\phi)_{[-1]}))
\end{equation}
This architecture implicitly learns to measure the ``competence'' of the base model weights $W_\theta$ for the given input $X$ by learning the necessary correction $W_\phi$ required to map the state to its value estimate.

\subsection{Linear Probe}
The Linear Probe shares the same paradigm as the proposed probing methods. It collects hidden states from the base model and applies linear regression to each layer to obtain a predicted value. The average of these predicted values is then used to estimate the probability of correctness.

\subsection{Loss Curve Fit}
Inspired by \citet{zai-loss}, who utilized regression to link loss values to downstream performance, we model the relationship between the model's loss on a given input and its final probability of correctness. Specifically, we collect loss values and the corresponding correctness probabilities for all prompts and perform linear regression using ordinary least squares. To predict downstream performance, we input the loss of a given prompt into the trained regression model to estimate its probability of correctness.

\section{Full Results}
\label{app:results}
In this section, we present the full results on each dataset for checkpoints from the pre-training stage (base model) and the post-training stage (instruct and think).

\subsection{Base Model}
\begin{table*}[htbp]
  \centering
  \caption{Comprehensive AUROC Results for OLMo-3-base across Six Datasets (Train ckpt $\times$ Test Ckpt)}
  \label{tab:olmo3-all-datasets}

  \begin{subtable}{0.49\linewidth}
    \centering
    \caption{MMLU Dataset}
    \resizebox{\linewidth}{!}{
      \begin{tabular}{c|c|c|c|c|c|c|c|c}
        \toprule
        \multirow{2}{*}{\makecell{Train\\Ckpt (k)}} & \multicolumn{8}{c}{Test Ckpt (k)}\\
        \cline{2-9}
        & 200 & 400 & 600 & 800 & 1000 & 1200 & 1400 & last \\
        \midrule
        200  & 0.6613 & 0.6739 & 0.6748 & 0.6804 & 0.7062 & 0.6875 & 0.6877 & 0.6984\\ 
        400  & --     & 0.6685 & 0.6734 & 0.6838 & 0.7119 & 0.6884 & 0.6909 & 0.6997\\ 
        600  & --     & --     & 0.6750 & 0.6821 & 0.7087 & 0.6849 & 0.6878 & 0.6945\\ 
        800  & --     & --     & --     & 0.6737 & 0.7141 & 0.6969 & 0.7004 & 0.7072\\ 
        1000 & --     & --     & --     & --     & 0.7230 & 0.7014 & 0.7062 & 0.7049\\ 
        1200 & --     & --     & --     & --     & --     & 0.6939 & 0.6967 & 0.6950\\ 
        1400 & --     & --     & --     & --     & --     & --     & 0.6987 & 0.7018\\
        \bottomrule
      \end{tabular}
    }
  \end{subtable}
  \hfill 
  \begin{subtable}{0.49\linewidth}
    \centering
    \caption{GSM8k Dataset}
    \resizebox{\linewidth}{!}{
      \begin{tabular}{c|c|c|c|c|c|c|c|c}
        \toprule
        \multirow{2}{*}{\makecell{Train\\Ckpt (k)}} & \multicolumn{8}{c}{Test Ckpt (k)}\\
        \cline{2-9}
        & 200 & 400 & 600 & 800 & 1000 & 1200 & 1400 & last \\
        \midrule
        200  & 0.8082 & 0.7463 & 0.7279 & 0.7376 & 0.7549 & 0.7112 & 0.7253 & 0.7439\\ 
        400  & --     & 0.7742 & 0.7592 & 0.7594 & 0.7786 & 0.7336 & 0.7374 & 0.7648\\ 
        600  & --     & --     & 0.7771 & 0.7654 & 0.7825 & 0.7453 & 0.7489 & 0.7669\\ 
        800  & --     & --     & --     & 0.7827 & 0.7781 & 0.7548 & 0.7522 & 0.7748\\ 
        1000 & --     & --     & --     & --     & 0.7732 & 0.7399 & 0.7478 & 0.7568\\ 
        1200 & --     & --     & --     & --     & --     & 0.7564 & 0.7490 & 0.7671\\ 
        1400 & --     & --     & --     & --     & --     & --     & 0.7554 & 0.7670\\
        \bottomrule
      \end{tabular}
    }
  \end{subtable}

  \par\vspace{1em} 

  \begin{subtable}{0.49\linewidth}
    \centering
    \caption{MATH Dataset}
    \resizebox{\linewidth}{!}{
      \begin{tabular}{c|c|c|c|c|c|c|c|c}
        \toprule
        \multirow{2}{*}{\makecell{Train\\Ckpt (k)}} & \multicolumn{8}{c}{Test Ckpt (k)}\\
        \cline{2-9}
        & 200 & 400 & 600 & 800 & 1000 & 1200 & 1400 & last \\
        \midrule
        200  & 0.7096 & 0.6997 & 0.7525 & 0.7662 & 0.7944 & 0.7490 & 0.7158 & 0.7224\\ 
        400  & --     & 0.6863 & 0.7576 & 0.7768 & 0.8130 & 0.7544 & 0.7326 & 0.7361\\ 
        600  & --     & --     & 0.7520 & 0.7695 & 0.8072 & 0.7484 & 0.7294 & 0.7283\\ 
        800  & --     & --     & --     & 0.7776 & 0.7966 & 0.7562 & 0.7289 & 0.7328\\ 
        1000 & --     & --     & --     & --     & 0.8236 & 0.7628 & 0.7322 & 0.7308\\ 
        1200 & --     & --     & --     & --     & --     & 0.7476 & 0.7263 & 0.7303\\ 
        1400 & --     & --     & --     & --     & --     & --     & 0.7347 & 0.7538\\
        \bottomrule
      \end{tabular}
    }
  \end{subtable}
  \hfill
  \begin{subtable}{0.49\linewidth}
    \centering
    \caption{BBH Dataset}
    \resizebox{\linewidth}{!}{
      \begin{tabular}{c|c|c|c|c|c|c|c|c}
        \toprule
        \multirow{2}{*}{\makecell{Train\\Ckpt (k)}} & \multicolumn{8}{c}{Test Ckpt (k)}\\
        \cline{2-9}
        & 200 & 400 & 600 & 800 & 1000 & 1200 & 1400 & last \\
        \midrule
        200  & 0.7749 & 0.7583 & 0.7782 & 0.7773 & 0.7546 & 0.7633 & 0.7885 & 0.7927\\ 
        400  & --     & 0.7709 & 0.7778 & 0.7649 & 0.7565 & 0.7691 & 0.7721 & 0.7804\\ 
        600  & --     & --     & 0.8046 & 0.7853 & 0.7625 & 0.7839 & 0.7827 & 0.7844\\ 
        800  & --     & --     & --     & 0.8014 & 0.7765 & 0.7961 & 0.8020 & 0.8024\\ 
        1000 & --     & --     & --     & --     & 0.7719 & 0.7732 & 0.7929 & 0.7933\\ 
        1200 & --     & --     & --     & --     & --     & 0.7982 & 0.7965 & 0.7993\\ 
        1400 & --     & --     & --     & --     & --     & --     & 0.8046 & 0.7972\\
        \bottomrule
      \end{tabular}
    }
  \end{subtable}

  \par\vspace{1em} 

  \begin{subtable}{0.49\linewidth}
    \centering
    \caption{HumanEval Dataset}
    \resizebox{\linewidth}{!}{
      \begin{tabular}{c|c|c|c|c|c|c|c|c}
        \toprule
        \multirow{2}{*}{\makecell{Train\\Ckpt (k)}} & \multicolumn{8}{c}{Test Ckpt (k)}\\
        \cline{2-9}
        & 200 & 400 & 600 & 800 & 1000 & 1200 & 1400 & last \\
        \midrule
        200  & 0.9252 & 0.8483 & 0.8667 & 0.9197 & 0.7836 & 0.8298 & 0.8496 & 0.8748\\ 
        400  & --     & 0.8594 & 0.8913 & 0.9218 & 0.8012 & 0.8655 & 0.8871 & 0.9093\\ 
        600  & --     & --     & 0.8952 & 0.9340 & 0.8064 & 0.8626 & 0.8955 & 0.9174\\ 
        800  & --     & --     & --     & 0.9163 & 0.8211 & 0.8538 & 0.8768 & 0.8905\\ 
        1000 & --     & --     & --     & --     & 0.7965 & 0.8485 & 0.8732 & 0.8967\\ 
        1200 & --     & --     & --     & --     & --     & 0.8456 & 0.8720 & 0.8967\\ 
        1400 & --     & --     & --     & --     & --     & --     & 0.8841 & 0.9105\\
        \bottomrule
      \end{tabular}
    }
  \end{subtable}
  \hfill
  \begin{subtable}{0.49\linewidth}
    \centering
    \caption{MBPP Dataset}
    \resizebox{\linewidth}{!}{
      \begin{tabular}{c|c|c|c|c|c|c|c|c}
        \toprule
        \multirow{2}{*}{\makecell{Train\\Ckpt (k)}} & \multicolumn{8}{c}{Test Ckpt (k)}\\
        \cline{2-9}
        & 200 & 400 & 600 & 800 & 1000 & 1200 & 1400 & last \\
        \midrule
        200  & 0.7043 & 0.6809 & 0.6938 & 0.6986 & 0.7086 & 0.6862 & 0.6812 & 0.6758\\ 
        400  & --     & 0.7353 & 0.7331 & 0.7514 & 0.7270 & 0.7257 & 0.7122 & 0.7040\\ 
        600  & --     & --     & 0.7598 & 0.7738 & 0.7510 & 0.7449 & 0.7323 & 0.7197\\ 
        800  & --     & --     & --     & 0.7822 & 0.7450 & 0.7590 & 0.7478 & 0.7310\\ 
        1000 & --     & --     & --     & --     & 0.7485 & 0.7390 & 0.7256 & 0.7294\\ 
        1200 & --     & --     & --     & --     & --     & 0.7583 & 0.7524 & 0.7563\\ 
        1400 & --     & --     & --     & --     & --     & --     & 0.7477 & 0.7377\\
        \bottomrule
      \end{tabular}
    }
  \end{subtable}

\end{table*}

\newpage

\subsection{Instruct Model}
\begin{table*}[htbp]
  \centering
  \caption{OLMo-3-Instruct Results}
  \label{tab:3x3-grid}

  \begin{subtable}{0.32\linewidth}
    \centering
    \caption{Main Results} 
    \resizebox{\linewidth}{!}{
      \begin{tabular}{c|c|c|c|c}
        \toprule
        \multirow{2}{*}{\makecell{Train\\Ckpt}} & \multicolumn{4}{c}{Test Ckpt}\\
        \cline{2-5}
        & 100 & 200 & 300 & 400 \\
        \midrule
        100 & 0.7402 & 0.7261 & 0.7253 & 0.7421 \\
        200 & {--} & 0.7367 & 0.7379 & 0.7508 \\
        300 & {--} & {--} & 0.7285 & 0.7452 \\
        400 & {--} & {--} & {--} & 0.7293 \\
        \bottomrule
      \end{tabular}
    }
  \end{subtable}
  \hfill 
  \begin{subtable}{0.32\linewidth}
    \centering
    \caption{MMLU Results} 
    \resizebox{\linewidth}{!}{
      \begin{tabular}{c|c|c|c|c}
        \toprule
        \multirow{2}{*}{\makecell{Train\\Ckpt}} & \multicolumn{4}{c}{Test Ckpt}\\
        \cline{2-5}
        & 100 & 200 & 300 & 400 \\
        \midrule
        100 & 0.6706 & 0.6723 & 0.6670 & 0.6667 \\
        200 & {--} & 0.6674 & 0.6630 & 0.6631 \\
        300 & {--} & {--} & 0.6651 & 0.6655 \\
        400 & {--} & {--} & {--} & 0.6684 \\
        \bottomrule
      \end{tabular}
    }
  \end{subtable}
  \hfill
  \begin{subtable}{0.32\linewidth}
    \centering
    \caption{GSM8k Results} 
    \resizebox{\linewidth}{!}{
      \begin{tabular}{c|c|c|c|c}
        \toprule
        \multirow{2}{*}{\makecell{Train\\Ckpt}} & \multicolumn{4}{c}{Test Ckpt}\\
        \cline{2-5}
        & 100 & 200 & 300 & 400 \\
        \midrule
        100 & 0.7215 & 0.7267 & 0.7220 & 0.7264 \\
        200 & {--} & 0.7310 & 0.7238 & 0.7294 \\
        300 & {--} & {--} & 0.7200 & 0.7264 \\
        400 & {--} & {--} & {--} & 0.7164 \\
        \bottomrule
      \end{tabular}
    }
  \end{subtable}

  \vspace{1em} 

  \begin{subtable}{0.32\linewidth}
    \centering
    \caption{MATH Results}
    \resizebox{\linewidth}{!}{
      \begin{tabular}{c|c|c|c|c}
        \toprule
        \multirow{2}{*}{\makecell{Train\\Ckpt}} & \multicolumn{4}{c}{Test Ckpt}\\
        \cline{2-5}
        & 100 & 200 & 300 & 400 \\
        \midrule
        100 & 0.8365 & 0.8416 & 0.8172 & 0.8551 \\
        200 & {--} & 0.8527 & 0.8354 & 0.8656 \\
        300 & {--} & {--} & 0.8288 & 0.8665 \\
        400 & {--} & {--} & {--} & 0.8606 \\
        \bottomrule
      \end{tabular}
    }
  \end{subtable}
  \hfill
  \begin{subtable}{0.32\linewidth}
    \centering
    \caption{BBH Results}
    \resizebox{\linewidth}{!}{
      \begin{tabular}{c|c|c|c|c}
        \toprule
        \multirow{2}{*}{\makecell{Train\\Ckpt}} & \multicolumn{4}{c}{Test Ckpt}\\
        \cline{2-5}
        & 100 & 200 & 300 & 400 \\
        \midrule
        100 & 0.8856 & 0.8749 & 0.8721 & 0.8566 \\
        200 & {--} & 0.8825 & 0.8852 & 0.8702 \\
        300 & {--} & {--} & 0.8878 & 0.8777 \\
        400 & {--} & {--} & {--} & 0.8816 \\
        \bottomrule
      \end{tabular}
    }
  \end{subtable}
  \hfill
  \begin{subtable}{0.32\linewidth}
    \centering
    \caption{HumanEval Results}
    \resizebox{\linewidth}{!}{
      \begin{tabular}{c|c|c|c|c}
        \toprule
        \multirow{2}{*}{\makecell{Train\\Ckpt}} & \multicolumn{4}{c}{Test Ckpt}\\
        \cline{2-5}
        & 100 & 200 & 300 & 400 \\
        \midrule
        100 & 0.6215 & 0.6512 & 0.6861 & 0.6922 \\
        200 & {--} & 0.6887 & 0.7126 & 0.7172 \\
        300 & {--} & {--} & 0.7381 & 0.7188 \\
        400 & {--} & {--} & {--} & 0.6266 \\
        \bottomrule
      \end{tabular}
    }
  \end{subtable}

  \vspace{1em} 

  \begin{subtable}{0.32\linewidth}
    \centering
    \caption{MBPP Results}
    \resizebox{\linewidth}{!}{
      \begin{tabular}{c|c|c|c|c}
        \toprule
        \multirow{2}{*}{\makecell{Train\\Ckpt}} & \multicolumn{4}{c}{Test Ckpt}\\
        \cline{2-5}
        & 100 & 200 & 300 & 400 \\
        \midrule
        100 & 0.7999 & 0.7718 & 0.7652 & 0.7919 \\
        200 & {--} & 0.7694 & 0.7755 & 0.7961 \\
        300 & {--} & {--} & 0.7208 & 0.7553 \\
        400 & {--} & {--} & {--} & 0.7545 \\
        \bottomrule
      \end{tabular}
    }
  \end{subtable}
  \hfill
  \begin{subtable}{0.32\linewidth}
    \centering
    \caption{AIME Results}
    \resizebox{\linewidth}{!}{
      \begin{tabular}{c|c|c|c|c}
        \toprule
        \multirow{2}{*}{\makecell{Train\\Ckpt}} & \multicolumn{4}{c}{Test Ckpt}\\
        \cline{2-5}
        & 100 & 200 & 300 & 400 \\
        \midrule
        100 & 0.7267 & 0.7348 & 0.6244 & 0.7094 \\
        200 & {--} & 0.7593 & 0.6591 & 0.7302 \\
        300 & {--} & {--} & 0.6196 & 0.7179 \\
        400 & {--} & {--} & {--} & 0.7063 \\
        \bottomrule
      \end{tabular}
    }
  \end{subtable}
  \hfill
  \begin{subtable}{0.32\linewidth}
    \centering
    \caption{GPQA Results}
    \resizebox{\linewidth}{!}{
      \begin{tabular}{c|c|c|c|c}
        \toprule
        \multirow{2}{*}{\makecell{Train\\Ckpt}} & \multicolumn{4}{c}{Test Ckpt}\\
        \cline{2-5}
        & 100 & 200 & 300 & 400 \\
        \midrule
        100 & 0.6594 & 0.5353 & 0.6484 & 0.6385 \\
        200 & {--} & 0.5428 & 0.6486 & 0.6342 \\
        300 & {--} & {--} & 0.6480 & 0.6336 \\
        400 & {--} & {--} & {--} & 0.6199 \\
        \bottomrule
      \end{tabular}
    }
  \end{subtable}

\end{table*}

\newpage
\subsection{Think Model}
\begin{table*}[htbp]
  \centering
  \caption{OLMo-3-Think AUROC Results (Total 9 Datasets)}
  \label{tab:olmo3-2x5-layout}

  \begin{subtable}{0.48\linewidth}
    \centering
    \caption{Main Results}
    \resizebox{\linewidth}{!}{
      \begin{tabular}{c|cccccc}
        \toprule
        \multirow{2}{*}{\makecell{Train\\Ckpt}} & \multicolumn{6}{c}{Test Ckpt}\\
        \cline{2-7}
        & 200 & 400 & 600 & 800 & 1000 & 1200 \\
        \midrule
        200  & 0.7946 & 0.7922 & 0.7849 & 0.7969 & 0.8001 & 0.7872 \\
        400  & {--} & 0.7925 & 0.7934 & 0.7977 & 0.8030 & 0.7859 \\
        600  & {--} & {--} & 0.7917 & 0.8055 & 0.8108 & 0.7941 \\
        800  & {--} & {--} & {--} & 0.7967 & 0.7992 & 0.7862 \\
        1000 & {--} & {--} & {--} & {--} & 0.8126 & 0.7941 \\
        1200 & {--} & {--} & {--} & {--} & {--} & 0.7780 \\
        \bottomrule
      \end{tabular}
    }
  \end{subtable}
  
  \par\vspace{0.5em} 

  \begin{subtable}{0.48\linewidth}
    \centering
    \caption{MMLU Results}
    \resizebox{\linewidth}{!}{
      \begin{tabular}{c|cccccc}
        \toprule
        \multirow{2}{*}{\makecell{Train\\Ckpt}} & \multicolumn{6}{c}{Test Ckpt}\\ \cline{2-7}
        & 200 & 400 & 600 & 800 & 1000 & 1200 \\ \midrule
        200  & 0.6573 & 0.6558 & 0.6556 & 0.6493 & 0.6538 & 0.6541 \\ 
        400  & {--} & 0.6626 & 0.6621 & 0.6556 & 0.6600 & 0.6602 \\
        600  & {--} & {--} & 0.6683 & 0.6638 & 0.6686 & 0.6679 \\ 
        800  & {--} & {--} & {--} & 0.6578 & 0.6630 & 0.6630 \\
        1000 & {--} & {--} & {--} & {--} & 0.6669 & 0.6664 \\ 
        1200 & {--} & {--} & {--} & {--} & {--} & 0.6649 \\ 
        \bottomrule
      \end{tabular}
    }
  \end{subtable}
  \hfill
  \begin{subtable}{0.48\linewidth}
    \centering
    \caption{GSM8k Results}
    \resizebox{\linewidth}{!}{
      \begin{tabular}{c|cccccc}
        \toprule
        \multirow{2}{*}{\makecell{Train\\Ckpt}} & \multicolumn{6}{c}{Test Ckpt}\\ \cline{2-7}
        & 200 & 400 & 600 & 800 & 1000 & 1200 \\ \midrule
        200  & 0.7578 & 0.7632 & 0.7408 & 0.7424 & 0.7635 & 0.7540 \\ 
        400  & {--} & 0.7581 & 0.7417 & 0.7416 & 0.7642 & 0.7520 \\ 
        600  & {--} & {--} & 0.7594 & 0.7588 & 0.7801 & 0.7591 \\ 
        800  & {--} & {--} & {--} & 0.7481 & 0.7703 & 0.7564 \\ 
        1000 & {--} & {--} & {--} & {--} & 0.7681 & 0.7465 \\ 
        1200 & {--} & {--} & {--} & {--} & {--} & 0.7430 \\ 
        \bottomrule
      \end{tabular}
    }
  \end{subtable}

  \par\vspace{0.5em} 

  \begin{subtable}{0.48\linewidth}
    \centering
    \caption{MATH Results}
    \resizebox{\linewidth}{!}{
      \begin{tabular}{c|cccccc}
        \toprule
        \multirow{2}{*}{\makecell{Train\\Ckpt}} & \multicolumn{6}{c}{Test Ckpt}\\ \cline{2-7}
        & 200 & 400 & 600 & 800 & 1000 & 1200 \\ \midrule
        200  & 0.9299 & 0.9129 & 0.8870 & 0.8938 & 0.8884 & 0.8667 \\ 
        400  & {--} & 0.9061 & 0.8816 & 0.8897 & 0.8806 & 0.8623 \\ 
        600  & {--} & {--} & 0.8812 & 0.8903 & 0.8866 & 0.8733 \\ 
        800  & {--} & {--} & {--} & 0.8853 & 0.8850 & 0.8728 \\ 
        1000 & {--} & {--} & {--} & {--} & 0.8917 & 0.8825 \\ 
        1200 & {--} & {--} & {--} & {--} & {--} & 0.8657 \\ 
        \bottomrule
      \end{tabular}
    }
  \end{subtable}
  \hfill
  \begin{subtable}{0.48\linewidth}
    \centering
    \caption{BBH Results}
    \resizebox{\linewidth}{!}{
      \begin{tabular}{c|cccccc}
        \toprule
        \multirow{2}{*}{\makecell{Train\\Ckpt}} & \multicolumn{6}{c}{Test Ckpt}\\ \cline{2-7}
        & 200 & 400 & 600 & 800 & 1000 & 1200 \\ \midrule
        200  & 0.9557 & 0.9542 & 0.9290 & 0.9514 & 0.9465 & 0.9303 \\
        400  & {--} & 0.9505 & 0.9258 & 0.9460 & 0.9475 & 0.9331 \\ 
        600  & {--} & {--} & 0.9188 & 0.9475 & 0.9476 & 0.9298 \\ 
        800  & {--} & {--} & {--} & 0.9573 & 0.9482 & 0.9317 \\ 
        1000 & {--} & {--} & {--} & {--} & 0.9533 & 0.9364 \\ 
        1200 & {--} & {--} & {--} & {--} & {--} & 0.9298 \\ \bottomrule
      \end{tabular}
    }
  \end{subtable}

  \par\vspace{0.5em} 

  \begin{subtable}{0.48\linewidth}
    \centering
    \caption{HumanEval Results}
    \resizebox{\linewidth}{!}{
      \begin{tabular}{c|cccccc}
        \toprule
        \multirow{2}{*}{\makecell{Train\\Ckpt}} & \multicolumn{6}{c}{Test Ckpt}\\ \cline{2-7}
        & 200 & 400 & 600 & 800 & 1000 & 1200 \\ \midrule
        200  & 0.7480 & 0.7434 & 0.7492 & 0.7990 & 0.8090 & 0.7902 \\ 
        400  & {--} & 0.7606 & 0.8041 & 0.8000 & 0.8343 & 0.7897 \\ 
        600  & {--} & {--} & 0.7646 & 0.8049 & 0.8194 & 0.7946 \\ 
        800  & {--} & {--} & {--} & 0.7926 & 0.8021 & 0.7788 \\ 
        1000 & {--} & {--} & {--} & {--} & 0.8462 & 0.8006 \\ 
        1200 & {--} & {--} & {--} & {--} & {--} & 0.7639 \\ 
        \bottomrule
      \end{tabular}
    }
  \end{subtable}
  \hfill
  \begin{subtable}{0.48\linewidth}
    \centering
    \caption{MBPP Results}
    \resizebox{\linewidth}{!}{
      \begin{tabular}{c|cccccc}
        \toprule
        \multirow{2}{*}{\makecell{Train\\Ckpt}} & \multicolumn{6}{c}{Test Ckpt}\\ \cline{2-7}
        & 200 & 400 & 600 & 800 & 1000 & 1200 \\ \midrule
        200  & 0.7191 & 0.7238 & 0.7476 & 0.7454 & 0.7394 & 0.7269 \\
        400  & {--} & 0.7171 & 0.7452 & 0.7535 & 0.7312 & 0.7180 \\ 
        600  & {--} & {--} & 0.7576 & 0.7679 & 0.7626 & 0.7297 \\ 
        800  & {--} & {--} & {--} & 0.7387 & 0.7270 & 0.7143 \\ 
        1000 & {--} & {--} & {--} & {--} & 0.7493 & 0.7325 \\ 
        1200 & {--} & {--} & {--} & {--} & {--} & 0.7005 \\ 
        \bottomrule
      \end{tabular}
    }
  \end{subtable}

  \par\vspace{0.5em} 
  \begin{subtable}{0.48\linewidth}
    \centering
    \caption{AIME Results}
    \resizebox{\linewidth}{!}{
      \begin{tabular}{c|cccccc}
        \toprule
        \multirow{2}{*}{\makecell{Train\\Ckpt}} & \multicolumn{6}{c}{Test Ckpt}\\ \cline{2-7}
        & 200 & 400 & 600 & 800 & 1000 & 1200 \\ \midrule
        200  & 0.6575 & 0.6903 & 0.7069 & 0.7008 & 0.6855 & 0.6722 \\
        400  & {--} & 0.6726 & 0.7028 & 0.6834 & 0.6584 & 0.6580 \\ 
        600  & {--} & {--} & 0.6992 & 0.6863 & 0.6697 & 0.6610 \\ 
        800  & {--} & {--} & {--} & 0.6887 & 0.6765 & 0.6657 \\ 
        1000 & {--} & {--} & {--} & {--} & 0.6663 & 0.6492 \\ 
        1200 & {--} & {--} & {--} & {--} & {--} & 0.6675 \\ 
        \bottomrule
      \end{tabular}
    }
  \end{subtable}
  \hfill
  \begin{subtable}{0.48\linewidth}
    \centering
    \caption{GPQA Results}
    \resizebox{\linewidth}{!}{
      \begin{tabular}{c|cccccc}
        \toprule
        \multirow{2}{*}{\makecell{Train\\Ckpt}} & \multicolumn{6}{c}{Test Ckpt}\\ \cline{2-7}
        & 200 & 400 & 600 & 800 & 1000 & 1200 \\ \midrule
        200  & 0.6808 & 0.6294 & 0.6412 & 0.6408 & 0.6509 & 0.6112 \\
        400  & {--} & 0.6417 & 0.6747 & 0.6551 & 0.6822 & 0.6287 \\
        600  & {--} & {--} & 0.6522 & 0.6614 & 0.6551 & 0.6120 \\ 
        800  & {--} & {--} & {--} & 0.6672 & 0.6709 & 0.6210 \\ 
        1000 & {--} & {--} & {--} & {--} & 0.6500 & 0.6058 \\
        1200 & {--} & {--} & {--} & {--} & {--} & 0.6222 \\ 
        \bottomrule
      \end{tabular}
    }
  \end{subtable}
  \vspace{-8em}
\end{table*}

\newpage
\subsection{Results on OLMo3-32B Base Model} \label{app:result-olmo3-32b}

\Cref{tab:olmo3-32b-merged} shows the results on OLMo3-32B based model\footnote{\url{https://huggingface.co/allenai/Olmo-3-1125-32B}} using submodel probe for prediction.

\begin{table*}[htbp]
  \centering
  \caption{OLMo-3-32B AUROC Results (all tested on checkpoint 656k).}
  \label{tab:olmo3-32b-merged}
  \resizebox{\textwidth}{!}{
    \begin{tabular}{c|ccccccc}
      \toprule
      \makecell{Train\\Ckpt (k)} & Overall & BBH & GSM8k & HumanEval & MBPP & MMLU & MATH \\
      \midrule
      100 & 0.7473 & 0.6928 & 0.7580 & 0.7885 & 0.7256 & 0.7062 & 0.8125 \\
      200 & 0.7556 & 0.7365 & 0.7608 & 0.7652 & 0.7512 & 0.7033 & 0.8169 \\
      300 & 0.7582 & 0.7474 & 0.7636 & 0.7637 & 0.7600 & 0.6982 & 0.8160 \\
      400 & 0.7605 & 0.7758 & 0.7641 & 0.7373 & 0.7562 & 0.7220 & 0.8078 \\
      500 & 0.7694 & 0.7832 & 0.7809 & 0.7803 & 0.7500 & 0.7071 & 0.8150 \\
      600 & 0.7573 & 0.7772 & 0.7770 & 0.7687 & 0.7162 & 0.7021 & 0.8027 \\
      \bottomrule
    \end{tabular}
  }
\end{table*}

\subsection{MSE Results on Apertus-8B} \label{app:apertus-mse}

\begin{table*}[htbp]
    \centering
    \caption{Apertus-8B-2509 MSE Results}
    \resizebox{0.92\linewidth}{!}{
    \begin{tabular}{c|c|c|c|c|c|c|c|c}
    \toprule
    & \multirow{2}{*}{\makecell{Train\\Ckpt (k)}} & \multicolumn{7}{c}{\textbf{Test Ckpt (k)}} \\
    \cline{3-9}
    & & \textbf{400} & \textbf{800} & \textbf{1194} & \textbf{1670} & \textbf{2000} & \textbf{2400} & \textbf{2627} \\
    \midrule
    \multirow{7}{*}{Submodel}
    & 400  & \textbf{0.0798} & 0.0819 & 0.0821 & 0.0922 & 0.1079 & 0.1232 & 0.1381 \\
    & 800  & -- & \textbf{0.0798} & 0.0833 & 0.0943 & 0.1057 & 0.1213 & 0.1452 \\
    & 1194 & -- & -- & \textbf{0.0820} & 0.0950 & 0.1096 & 0.1295 & 0.1603 \\
    & 1670 & -- & -- & -- & \textbf{0.0864} & 0.1004 & 0.1169 & 0.1361 \\
    & 2000 & -- & -- & -- & -- & \textbf{0.0953} & 0.1145 & 0.1349 \\
    & 2400 & -- & -- & -- & -- & -- & \textbf{0.1029} & 0.1186 \\
    & 2627 & -- & -- & -- & -- & -- & -- & \textbf{0.1093} \\
    \bottomrule
    \end{tabular}
    }
    \label{tab:apertus-mse}
\end{table*}

\newpage
\subsection{Discussion on Performance}
Comparing across datasets, HumanEval, BBH, MATH and MBPP are relatively easy to predict, while MMLU, AIME and GPQA are hard. 
AIME and GPQA are extremely hard benchmarks that require ultra long reasoning steps to provide the correct answer. Contrary to our initial expectations of poor performance, we were surprised to find that the probe exhibited meaningful predictive capability on these two datasets. Specifically, it generally achieved scores above 0.6 on both instruct model and think model, and even surpassed 0.75 on AIME in instruct model. This suggests a possibility that, with further research, datasets like AIME, which typically require tens of thousands of tokens for evaluation, could be assessed without generation, relying solely on the internal states elicited by the input prompt to predict correctness.
For MMLU, we attribute the subpar performance to insufficient training data. Although the full MMLU training set contains nearly 100k samples, we utilized only 10k (concerning the generation time). This sample size is relatively small compared to the $\sim$14k test samples, preventing the probe from fully capturing the dataset's features and ultimately resulting in mediocre predictive capability.

\section{Comparison with Subset Sampling} \label{app:subset_sampling}
Subset sampling is an efficient evaluation method commonly used in industry. It approximates an LLM's performance on a full dataset by evaluating it on a smaller subset. The comparison between probes and subset sampling primarily focuses on two aspects: 1. If both methods are constrained to the same level of efficiency, which one provides a more accurate assessment of the model's performance on the full dataset? 2. Without excessive efficiency constraints, can the probe method maintain its performance on the sampled subset, thereby consistently achieving higher efficiency than the subset approach?

For aspect 1, we constrain the number of evaluation steps to 50 iters. Therefore, to match the probe's efficiency, the time cost of subset sampling is $ 50 \times T_{subset}=T_{init}+50\times{t_eval}\geq T_{init}$, as $t_{eval}$ is negligible. Then, we can get that $T_{subset}=\frac{T_{init}}{50}=\frac{T_{fullset}+T_{train}}{50}\geq\frac{T_{fullset}}{50}$. Therefore, we set 2\% as the proportion of subset size. The evaluation results are shown in \Cref{tab:subset_sampling}. It demonstrates that compared with subset sampling, probe evaluation obtains an overall smaller absolute error in assessing the performance on the full dataset.

\begin{table}[htbp]
    \centering
    \caption{Comparison between probe evaluation and subset sampling on absolute error (smaller is better).}
    \begin{tabular}{l|c|c|c|c|c|c}
    \toprule
       Absolute Error	&HumanEval	&MATH	&BBH	&MMLU	&GSM8k&	MBPP\\
       \midrule
Probe	&0.0263	&0.0009&	0.0447	&0.0159	&0.0081&	0.0952\\
Subset	&0.2305&	0.0012	&0.1221	&0.0320	&0.009	&0.0761\\
\bottomrule
    \end{tabular}
    \label{tab:subset_sampling}
\end{table}

For aspect 2, we set the subset size to 20\%, and conduct a probe evaluation on the subset. \Cref{tab:subset_probe} shows the comparison between probe trained on subset and trained on fullset. It demonstrates that even when the subset size is reduced to 20\%, the probe's performance on the subset remains close to its performance on the fullset, with only a slight and minor decline.

\begin{table}[htbp]
    \centering
    \caption{Comparison between probe trained on subset and fullset.}
    \begin{tabular}{l|l|c|c|c|c|c|c}
\toprule
Metrics	&Datasets	&HumanEval	&Math	&BBH	&MMLU	&GSM8K	&MBPP\\
\midrule
\multirow{2}{*}{AUROC}	&Subset&	0.8946	&0.7766&	0.7373	&0.6824&0.7232	&0.6251\\
&Fullset	&0.9163	&0.7776	&0.8014	&0.6737	&0.7827&	0.7822\\
\midrule
\multirow{2}{*}{MSE}	&Subset	&0.1035&	0.0354	&0.1520&	0.1566&	0.1167	&0.1692\\
&Fullset	&0.0810	&0.0333	&0.1302	&0.1492&	0.0962&	0.1395\\
\bottomrule
    \end{tabular}
    \label{tab:subset_probe}
\end{table}

\section{Time Comparison}
\label{app:time}
In this section, we present the time costs for the two models in the post-training stage (Instruct and Think). For the Instruct model, probe evaluation achieves a 61.4x speedup compared to generative evaluation, with its amortized time cost falling below that of generative evaluation after 11 evaluation steps. For the Think model, the efficiency gain reaches 231.8x, becoming more cost-effective than generative evaluation after just 9 evaluation steps.

\begin{figure}[htbp]
  \centering
  \begin{subfigure}{0.48\linewidth}
    \centering
    \includegraphics[width=\linewidth]{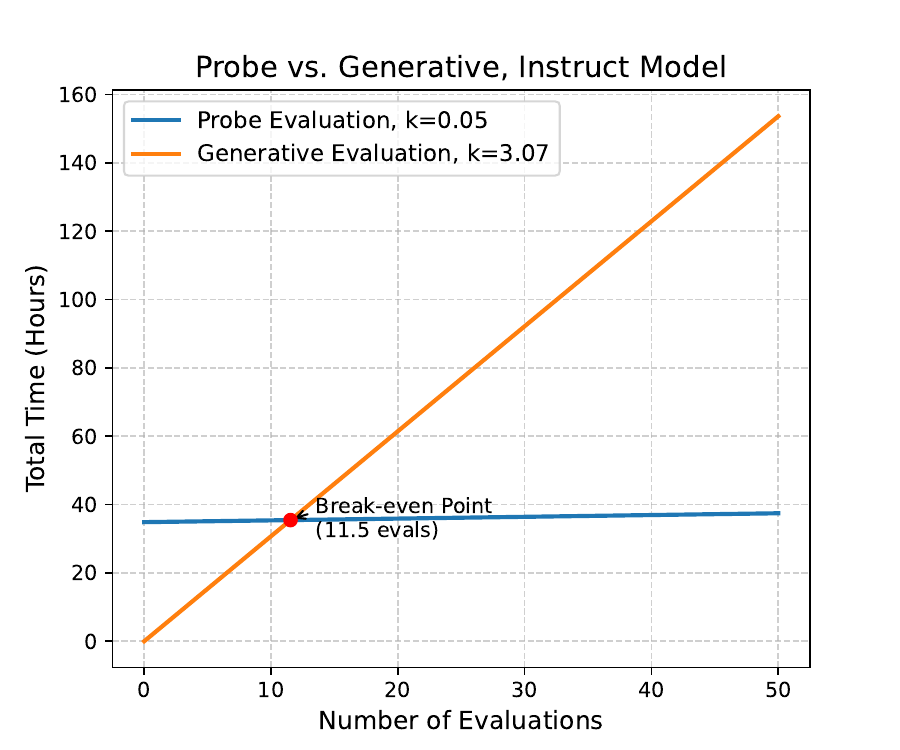}
    \caption{OLMo-3-7B-Instruct}
    \label{fig:instruction_time}
  \end{subfigure}
  \hfill
  \begin{subfigure}{0.48\linewidth}
    \centering
    \includegraphics[width=\linewidth]{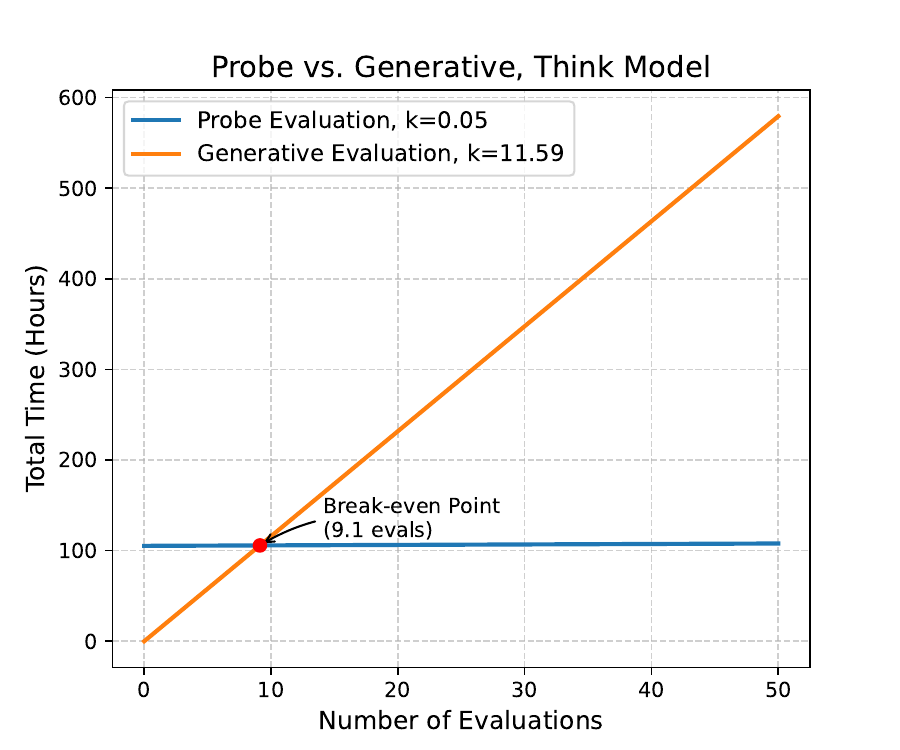}
    \caption{OLMo-3-7B-Think}
    \label{fig:think_time}
  \end{subfigure}
  
  \caption{Cumulative time consumption for Probe Evaluation vs. Generative Evaluation on post-training checkpoints for Instruct and Think models.}
  \label{fig:time_comparison}
\end{figure}

The probe preparation time includes the data collection time (i.e., conduct a generative evaluation on trainset) and the probe training time, as indicated by the y-axis intercept of the ``probe evaluation'' line in \Cref{fig:base_time} and \Cref{fig:time_comparison}. The specific time values are presented in \Cref{tab:preparation_time}.

\begin{table}[htbp]
    \centering
    \begin{tabular}{l|c|c}
    \toprule
         &  data collection & probe training\\
    \midrule
    base & 5.9h & 2.5h \\
    instruct & 35.5h & 2.6h\\
    think & 112h & 2.6h\\
    \bottomrule
    \end{tabular}
    \caption{Detailed time cost of probe preparation.}
    \label{tab:preparation_time}
\end{table}

\section{Discussion}

\subsection{Data Selection in Real-World LLM Training Monitoring}

Another discussion concerns the selection of training data. To ensure experimental rigor in this study, we maintained a strict separation between training and test sets, ensuring the test data remained unseen during the probe's training. This setup implicitly tests the probe's ability to generalize from training samples to test samples. However, in real-world industrial scenarios, this type of generalization is not strictly necessary. In practice, the primary objective is to accurately model the mapping from input data to downstream metrics; overfitting to the specific dataset is acceptable, provided that cross-checkpoint generalization is preserved. Consequently, practical applications can forgo the train/test split and train the probe directly on the target evaluation set. This approach would yield higher monitoring accuracy and further reduce the time overhead of the probe preparation phase.

\subsection{Limitation}

We have discussed the principles of using probes for LLM downstream performance monitoring and verify the feasibility of this approach. However, due to space constraints, we were unable to investigate how to further improve the predictive performance of the probes. This could be achieved by modifying the probe architecture or altering the input features, and future work will focus on further research in this direction.

\subsection{Potential Impact}

This work provides an efficient downstream performance monitoring tool for LLM training. However, we have not yet identified any potential negative impacts it might introduce.

\section{Hardware Specifications}

Key specifications relevant to the GPU used in experiments are detailed below:
\begin{itemize}[leftmargin=*, itemsep=0em, topsep=0.em]
\item Memory Capacity: 48 GB GDDR6
\item Memory Bandwidth: 960 GB/s
\item CUDA Cores: 14,080
\item Tensor Cores: 440
\item Single-Precision Performance: 69.3 TFLOPS
\item Tensor Performance: 1,108.4 TFLOPS
\item Interface: PCIe 4.0 $\times$16
\end{itemize}



\newpage

\section*{NeurIPS Paper Checklist}

\begin{enumerate}

\item {\bf Claims}
    \item[] Question: Do the main claims made in the abstract and introduction accurately reflect the paper's contributions and scope?
    \item[] Answer: \answerYes{} 
    \item[] Justification: The abstract and introduction clearly state the claims about proposing an in-training probe evaluation framework and validate its accuracy and efficiency against standard generative methods.

\item {\bf Limitations}
    \item[] Question: Does the paper discuss the limitations of the work performed by the authors?
    \item[] Answer: \answerYes{} 
    \item[] Justification: The paper includes a section discussing the limitation in further improving the predictive performance of the probes.

\item {\bf Theory assumptions and proofs}
    \item[] Question: For each theoretical result, does the paper provide the full set of assumptions and a complete (and correct) proof?
    \item[] Answer: \answerNA{} 
    \item[] Justification: The paper focuses on empirical evaluation, probe architectures, and modeling LLM performance, and does not include formal theoretical proofs or theorems.

    \item {\bf Experimental result reproducibility}
    \item[] Question: Does the paper fully disclose all the information needed to reproduce the main experimental results of the paper to the extent that it affects the main claims and/or conclusions of the paper (regardless of whether the code and data are provided or not)?
    \item[] Answer: \answerYes{} 
    \item[] Justification: The paper provides detailed experimental setups, including dataset configurations, model choices, evaluation metrics, and specific prompts used in Appendix.

\item {\bf Open access to data and code}
    \item[] Question: Does the paper provide open access to the data and code, with sufficient instructions to faithfully reproduce the main experimental results, as described in supplemental material?
    \item[] Answer: \answerYes{} 
    \item[] Justification: The paper utilizes open-source models and public datasets, and provides url source of these models. Code is available in supplement materials.

\item {\bf Experimental setting/details}
    \item[] Question: Does the paper specify all the training and test details (e.g., data splits, hyperparameters, how they were chosen, type of optimizer) necessary to understand the results?
    \item[] Answer: \answerYes{} 
    \item[] Justification: The experimental settings, such as data preparation, subsets used, and specific prompt formats for each task, are detailed in Section 4.1 and extensively in Appendix A.

\item {\bf Experiment statistical significance}
    \item[] Question: Does the paper report error bars suitably and correctly defined or other appropriate information about the statistical significance of the experiments?
    \item[] Answer: \answerNo{} 
    \item[] Justification: The paper reports aggregated evaluation metrics, which does not require error bar or statistical analysis.

\item {\bf Experiments compute resources}
    \item[] Question: For each experiment, does the paper provide sufficient information on the computer resources (type of compute workers, memory, time of execution) needed to reproduce the experiments?
    \item[] Answer: \answerYes{} 
    \item[] Justification: The paper explicitly details the hardware specifications used for the experiments in Appendix, including GPU memory, bandwidth, CUDA cores, and TFLOPS.
    
\item {\bf Code of ethics}
    \item[] Question: Does the research conducted in the paper conform, in every respect, with the NeurIPS Code of Ethics \url{https://neurips.cc/public/EthicsGuidelines}?
    \item[] Answer: \answerYes{} 
    \item[] Justification: The research involves standard LLM benchmarking and modeling on publicly available data and models, posing no apparent conflict with the NeurIPS Code of Ethics.

\item {\bf Broader impacts}
    \item[] Question: Does the paper discuss both potential positive societal impacts and negative societal impacts of the work performed?
    \item[] Answer: \answerYes{} 
    \item[] Justification: The paper includes a discussion about potential impact in appendix.
    
\item {\bf Safeguards}
    \item[] Question: Does the paper describe safeguards that have been put in place for responsible release of data or models that have a high risk for misuse (e.g., pre-trained language models, image generators, or scraped datasets)?
    \item[] Answer: \answerNA{} 
    \item[] Justification: The paper does not release new models or scraped datasets.

\item {\bf Licenses for existing assets}
    \item[] Question: Are the creators or original owners of assets (e.g., code, data, models), used in the paper, properly credited and are the license and terms of use explicitly mentioned and properly respected?
    \item[] Answer: \answerYes{} 
    \item[] Justification: The paper properly cites all the existing datasets (e.g., MMLU, GSM8K, MATH) and the base models (OLMo-3) utilized in the research.

\item {\bf New assets}
    \item[] Question: Are new assets introduced in the paper well documented and is the documentation provided alongside the assets?
    \item[] Answer: \answerNA{} 
    \item[] Justification: The paper utilizes existing public datasets and models and does not introduce or release any new assets.

\item {\bf Crowdsourcing and research with human subjects}
    \item[] Question: For crowdsourcing experiments and research with human subjects, does the paper include the full text of instructions given to participants and screenshots, if applicable, as well as details about compensation (if any)? 
    \item[] Answer: \answerNA{} 
    \item[] Justification: The research relies solely on automated generative evaluation and probe evaluation of language models and does not involve human subjects or crowdsourcing.

\item {\bf Institutional review board (IRB) approvals or equivalent for research with human subjects}
    \item[] Question: Does the paper describe potential risks incurred by study participants, whether such risks were disclosed to the subjects, and whether Institutional Review Board (IRB) approvals (or an equivalent approval/review based on the requirements of your country or institution) were obtained?
    \item[] Answer: \answerNA{} 
    \item[] Justification: The research does not involve human subjects.

\item {\bf Declaration of LLM usage}
    \item[] Question: Does the paper describe the usage of LLMs if it is an important, original, or non-standard component of the core methods in this research? Note that if the LLM is used only for writing, editing, or formatting purposes and does \emph{not} impact the core methodology, scientific rigor, or originality of the research, declaration is not required.
    \item[] Answer: \answerYes{} 
    \item[] Justification: The core methodology involves using LLMs (specifically OLMo3-7B, OLMo3-32B, and Apertus-8B checkpoints) and analyzing their internal representations.

\end{enumerate}

\end{document}